# A Foundation for Perception Computing, Logic, and Automata

Mohamed A. Belal
Associate professor
Faculty of Computers & Information
Helwan University, Egypt
dr.belal@alzaytoonah.edu.jo

## Abstract

In this report, a novel approach to intelligence and learning is introduced; this approach is based upon what we called perception logic. What we call '*perception automata*' is introduced in which learning is accomplished at different perception resolution. Learning in this automata is not heuristic, rather it guarantees the convergence of the approximated function to whatever precision required. Furthermore, the learning process can take place on-line and in at most O(log(N)) epochs, where N is the number of samples. The perception automata is based on hierarchal levels of resolution in which each level adds some details to the constructed function till the final level can successfully reconstruct the whole function. This approach combines the favors of computational approach in the sense that it is precise, structural and rigorous, and the features of distributed processing and adaptivity of soft computing, as well as continuity and real-time response of dynamical systems.

## 1. Introduction

In general, traditional A.I. techniques have not led to significantly better understanding of the perception process and pattern recognition. Till now, there is no unified approach to tackle the problem of intelligence engineering, Luger (pg. 803) [15] asked "Is it possible to give a formal, computational account of the process that enable intelligence?". Intelligence is viewed, by some scientists, as an abstract and immeasurable quantity. The main purpose of artificial intelligence is find a way of organizing knowledge in such a way that helps decision making to take place rapidly and efficiently; which means that learning should happen at optimal speed and in compact storage. Learning should also be able to discover relations between symbols that represent abstract concepts in outer world. Fundamentally, intelligence is not knowing *that* but rather knowing *how* [15]. Levesque & Brachman (1985) [14] have argued that intelligence needs less expressive representation and more efficient response or 'computation'. Others saw that intelligent action requires *a physical embodiment and integration* of an agent into the world [22,11].

The human brain can cope with the surrounding world, which is imprecise, uncertain, and adaptive [27]; this coping is not done by searching for an exact mathematical model to represent the nature, rather it is based on learning and adaptation. Therefore, soft computing principles were guided by these facts. Some theories interpret intelligence [15] as the ability to act and respond rather than trying to explains these actions, Husserl (70,72) [8] stated that "intelligence is not knowing what is true, rather, knowing how to cope in a world that is constantly changing and evolving" [15]. As an example, Animals recognize patterns without any ability to define them. Hence, recognition without definition characterizes much of intelligent behavior, it enables systems to generalize. Therefore, the information processing of human brain is not expected to be measurable, quantitative, and computational, rather the main function of the brain is to ensure survival in an adaptive and time varying environment [27].

Actually, the real world problems that our brain is required to perceive is high dimensional, non-linear, non-parametric, and with unknown dependency of variables [26]. This world is not a binary world; there are many states between zero and one, yes and no, black and white, and so on, there is a gradual change between these two extremes [27]. The mathematical model of computation, i.e. the Turing machine, is based on symbols, hence it deals with precise and well-defined symbols in order to start processing, the human brain structure does not resemble this model; it seems hard for it to convert every sensory information into distinct symbol, however, it has impressive ability to sense and act in responding to the outer world, this ability enforces us to think that our nervous system perform a very accurate and precise sort of computation that is neither symbol-based, nor even fuzzy-logic based, this is in spite of our impression to *measure* (not to take decision).

In computational systems, numbers are just symbols, and the formal logic yields its computation processes; when the computer performs arithmetic operations, it actually performs logical operations. However, can we say that the biological neural networks actually compute similar to this way? Boolean logic completely defines any computable function, the design of a computer system is called logic design, and this means that computation can be described in terms of that solid, formal, well-defined logic. On the other hand, we need another logic to describe our perception process, which should be also formal, well-defined, and unambiguous in which we contradict the "The physical symbol system hypothesis" of Newel and Simon [21] and consider the analog nature of the sensory signals.

Fuzziness in human brain comes from our inability to give an exact word or symbol of all what we observe or try to measure. For example, assume that there is a man looking at an analog device pointing at a position in between distinct positions '3' and '4' and its actual value is 3.37, when this man is asked what is the exact measurement; he will not be able to give an accurate answer; so we can conclude that the fuzziness comes from our ability to measure not in the sensory signal itself. The fuzzy logic based systems approach for function approximation or soft computing has drawbacks: Firstly, a structured human solution to the problem from the beginning, secondly, the ability to learn and adapt are difficult and weak, lastly, the number of rules increases exponentially as we increase the number of inputs [27]. However, one favor of fuzzy logic approach is that it leads to a reduction in the complexity of knowledge by applying fewer rules compared to the classical approach. This reduction comes from the capability of the fuzzy system of performing interpolation between rules, this property is still reserved in our approach. Hybrid models, such as fuzzy neural networks, though they combines the structural knowledge representation of fuzzy systems and the learning ability of neural networks, is lacking the ability of self-adaptation of its structure, besides the heuristic nature of the learning process which does not guarantee their convergence.

We agree with Zadeh's analysis of perception [29] that the universe is divided into information granules and every object has a degree of belonging to each information granule, but we differ in the point that we argue that the construction of these granules is dynamic and is based on another logic, which we call *perception logic*, and we will try to explain how the interaction of these *granules* leads to a very precise and accurate computation in the human brain. We should differentiate between the ability of the human brain to produce very accurate response to an outer stimuli, which is apparently based on very accurate computation, and its inability to bring out a *symbol* for each stimuli it faces; this can be explained as that naming each stimuli consumes a lot of cognitive space whereas all what is required is to map this stimuli to a proper response without trying to *name* it or *give it a symbol*.

In this report, we argue that the existence of fuzziness in human perception is a result of incomplete *symbolizing* of an event; or the inability to give exact measure of phenomena, there is no fuzzy signal inside the human brain, just a degree of perception of a certain event. However, if human being focuses to distinguish it, as a separate event or granular, (s)he will be able to describe it in an exact word. Therefore, we need a mathematical model that can express this phenomenon; we will call it "multi-resolution perception analysis". In fuzzy logic, the prerequisite existence of 'symbols' is inherited in the fuzzy model; those symbols are the fuzzy sets themselves. In the proposed perception logic terms, these symbols or bases functions are fixed and not subject to topology re-ordering, and this confine the flexibility of the learning automata.

The objective of this report is to propose a perception model that acts rationally, instead of a computational model that can appear intelligent. In this work, we argue that biological neural network is not an alternative to, or even based on, the formal logic approach, but, rather, it is based on its own logic or what we call '*perception logic*'. Specifically, in this research, we deal with the problems of common sense reasoning that could contribute significantly to the solution of real world problems in robotics, computer vision, machine learning and speech recognition. This study is not intended to be theoretical; rather, it is presented in an engineering perspective. One of our motivations of this work is that human being and living organisms respond similarly for similar stimulus, this leads us to assume the existence of some functional dependency in natural behavior. People and living organisms learns from events that repeat similarly. Random events could never lead to learning. These similarity and continuity enables us to generalize, predict and learn.

## 2. A.I Approaches to Perception

The meaning of the word 'perception' includes the construction of knowledge from sensory information (pg.35) [13], the discovery of associations between low level knowledge, and the organization of knowledge in which adaptation and recall are feasible and reliable.

Modeling perception, or generally, cognition can be classified into traditional symbolic approach, dynamical approach [19], and soft computing (or somehow, similarly, natural computing) approach. The law of qualitative structure of Newell and Simon [21] holds that cognition is a kind of Turing-Machine (TM) computations [24]. However, TM is ill-suited to certain important applications and what is called 'natural computing' [3].

Obviously human brain is not another version of Universal Turing Machine (UTM), cognitivism and 'Good Old Fashion Artificial Intelligence' GOFAI have assumed that cognition is a form of computation [9]. Furthermore, it is clear that soft computing in general is still far from replicating our neural system, there are a lot of tasks that we simply do, like seeing, hearing, and recognizing, these actions are still difficult to replicate, as we simply do them, in soft computing [27].

Cognition was seen, by some scientists, like Newell and Simon, as computational process, by computation here, we mean the symbolic logical manipulations that Turing machine can perform. Others saw it as a dynamical system, but the dominance of traditional A.I. in 60's and 70's made this approach not considered too much [19]. However, recently, the dynamical approach has restored its attention [19] by the work of Thelen and Smith (1994) [23], they argued that symbolic processing and computation are not the correct approach to understand cognition. Moreover, Freeman & Skarda, 1990 [7] saw symbols and representation are harmful to understand cognition. Van Gelder and Port (1995) [25] have showed that computational view of cognition, as transformation from one static symbol to another, is wrong, and the right view should be the dynamic system theory [19].

The reason why the TM computation approach was dominating in the era of Good Old Fashioned Artificial Intelligence (GOFAI) is that it provides a notation of mechanism in terms of computation notations; the computation in the form of TM machines provides a mechanism capable of processing symbols. Similar concepts can be applied in the understanding the self-production of biological cells [19]. This broadband definition of mechanism could help explaining how a complex system like the brain works [19]. One of the main features of the computation approach is that it provides an abstract specification of the casual organization of a system [6].The GOFAI scientists took computers as the model for cognition. Since computers assume problems to be solved in formal and symbolic way, then cognition would be assumed to be modeled similarly [9]. For example, Brooks (1995) has called the SMPA or Sense, Model, Plan, Action approach." [9].

The computational approach to cognition focuses on the structure to illustrate the process of information processing, but it does not explain the dynamic change of the mental process. The computational approach explains the manipulation of explicit, static symbols. The real situation is that we have emergent and active symbols that interact and change in time [19]. On the other hand, dynamical approach focuses on change in cognition systems, and contributes to describe coupling among the brain, body, and the environment [19].

There are many reasons to argue that cognition is a form of TM computation, the Von Neumann architecture is quite different form the architecture of the brain. The brain does not have central control, random access memory, and serial processing [19].

Recently, the dynamical approach to cognition has retained its attention, as it is believed that it has sufficient power, scope, and cohesion to count as a strong approach to understanding cognition [24]. Dynamical modeling is based upon dynamical system theory and includes the terms of attractors, transients, stability, coupling, bifurcations, chaos and so forth [24]. **Hence,** the dynamical system is a set of quantitative variables changing over time in accordance with dynamical laws described by some set of equations [24]. The representation in dynamical approach is seen as transient and changing rather than static, context free, permanent units [24].

The dynamical models treat the process of decision making as one in which the variables evolve interactively through time, this can provide a better understanding of the cognitive process in dealing with real time control [24]. Dynamical approaches explain the behavior of a system geometrically in terms of “trajectories", “attractors", “bifurcations", and so on. Dynamical systems are useful in analyzing “self-organization”, “emergent behavior", “relative stability”, and “component coupling [19].

In fact, the picture of human brain gives rise to a dynamical system with numerous variables changing over time and continuously coupling and interaction with each other. This structure could be described of a set of differential equations with sensory signals as inputs, motor neurons as outputs, and internal neurons as internal states **[9]**. The dynamical approach to cognition is guided by the fact that the natural cognition is a dynamical phenomena and it is better to understand it in dynamical terms [24]. Dynamical theories of cognition are languages for describing continual temporal change in complex systems [19].

However, the problem of dynamical system approach to cognition is that it does not yield a concise single formation, but they share the characteristic of the continual coupling between the brain, body, and the environment in real time [19]. Hence, the dynamical approach focuses on change, whereas the computational approach focuses on the “internal structure” [19]. The problem of dynamical approach, pointed out by Clark [5], is of two folds: Scaling and style of explanation [19]. By scaling, it is meant that the dynamical system, in general, is extremely hard to be analyzed if it is high dimensional [19]. By style of explanation, we mean that the dynamical system analysis does not provide us with means to answer why the behavior of a given system appears to be like this, and how to compare and conclude results [19]. We do not expect the dynamical approach to catch up some aspects of cognition suitable for computational approach, such as story comprehension, but it seems suitable for significant range of cognitive phenomena [24].

The dynamical and the classical approach are not so opposed to each other. Since dynamical systems could perform computations on real numbers, such as analog computers, on the other hand classical approach can effectively simulate dynamical systems [24]. In computational approach, cognition is considered to happen over time. On the other hand, in dynamical approach the cognition happens in time [24]. In classic cognitive science, symbols and their manipulations are the basic building blocks. In dynamical approach, there are dynamical entities such system states and trajectories shaped by attractors positions [24]. Therefore, there is a need for a theory that combines both structure and change as a unified formation.

The soft computing approach provides another means to model perception or cognition, the favors of classical hard computing approach to artificial intelligence, are that they are precise, certain, and rigorous [27], however, one obstacle that faces symbolic computation is that it is not dynamical; symbols do not have derivatives, only smooth functions have [1]. On the other hand, soft computing approach is promising, since the real world, we live in, is not explicitly measured or even certain, and “Precision and certainty carry a cost” (Zadeh 1994) [28]. In contract to hard computing, soft computing is approximate, heuristic, informal, qualitative, and ad hoc solution [27]. However, in spite of the success of some soft computing methods, formal models still have their advantages of being quantitative, precise, and mathematically well defined. However, hard computing always stuck at problems that are complex, high dimensional, or NP complete. Therefore advances in soft computing hold potential for revolutionizing the field of A.I.

Natural computing, as suggested and defined by MacLennan [3] is a computation that occurs in nature or inspired by nature. Examples of natural computation include information processing in the brain, in the immune system, or in the evolution in nature [3]. Examples of computation inspired by nature include artificial neural nets, genetic algorithms, simulated immune systems, ant colony optimization, particle swarm optimization, and simulated annealing [3]. Applications such as autonomous robotics, real-time control systems, and distributed intelligent systems are most relevant to natural computing and should be understood in terms of it [3]. TM computation, as a computational model, can be used to mimic the behavior of biological systems. However, natural computation is real-time, flexible, adaptive, robust, and continuous [3]. Natural computing includes continuous and topological computation [3]. Continuity in computations allows smoothness over time and considers small changes [3]. Natural computation, such as that of neural networks, requires non-Turing models of computation [3]. Natural computing can be discrete or continuous [3], continuous computation could serve as an alternative model to the TM, and it is more relevant to natural computation in the brain. It is also more related to ANN, massively parallel analog machines [3].

In order to mimic natural cognition, we need a generalized model that is analogical and it can be reduced to digital model if only the two extremes are considered. A popular suggestion is to model it as a dynamical system with discrete attractors [10]. Natural or biological cognition is not based simply on propositional logic, actually neurons are much like sophisticated analog components, moreover sense organs, like the ear and the eye and effectors are analogical [10]. However, this does not mean that the brain can not deal with Boolean logic, it can easily distinguish between pairs of opposites: black and white, sweet and bitter, and so on [17].

An extended computation theory could be expected that provides a means to understand complex systems like the brain and biological systems [19]. "What we need is a rapprochement between computation and dynamics (between theories of structure and theories of change) that can provide both." [19]. Finally, instead of arguing whether the brain is a computer or not, one can say that the brain is able to implement various complex computations [6]. Can dynamics, computation, and adaptation come up with a unified approach? A question asked by Michel [19]. Basically, our nervous system seems to balance between continuous, "analog", perception-based computation and discrete, logic based computation [10]. Hybrid computation, as presented by [10], is a model that consists of both logic based and associative computing. This architecture can combine the best properties of the classic, logic-based digital computer, and associative, perceptually-based neural net computer [10].

It is more informative to say that any physical or biological computes whether the underlying technology is digital, analog, or even biological [3]. MacLennan [3] has argued that computation in its broader definition should include both digital (or discrete) and analog (or continuous) computation, and the practical success of digital computers has confined its meaning to Turing Machine-computation or TM computation. This argument is based primarily on the existence of differential analyzer, analog computers, and modern analog VLSI devices (Mead 1989 [18]). Hence, Turing machine does not represent all kinds of computations as defined by MacLennan [3], for example, analog computers represents another alternative of computation. A broadened definition of computation is needed in order to include both discrete and continuous computation. Obviously, continuous computation contradicts many of the assumption of TM model of computation, while it is more suitable for addressing the issues of natural computing [3]. Instead of trying to find TM model for nature computation, it may be better to introduce the term natural computing in its broader definition and understand the computation as it is in nature [3].

## 3. Learning Problems in Artificial Intelligence Systems

Learning can be modeled by traditional methods such as statistical, fuzzy, and approximation methods. However, learning from experimental data still needs a mathematical or analytical model [27].

Till now, there is no such a perfect intelligent system that can be completely said to 'learn', Luger [15] has declared "there have been only limited results with programs that in any interesting sense can be said to *learn* … Even fundamental issues such as organizing knowledge require further research". Learning process, in general, faces critical problems such as determining the sufficiency and cleanness of data [15].

The generalization or universal approximation of neural network, fuzzy logic matrix, and support vector machine is very interesting property that can be used to model highly nonlinear and partially known complex system, but there is always a fundamental problem in the prerequisites and computational techniques that is used in order to approximate a multi-variate function. Support vector machine is based on the statistical theory of learning; the probabilistic models always formalize the problem of randomness in data, whereas back- propagation networks is based on parameters estimation and approximation methods, fuzzy systems is based on fuzzy logic; which models the imprecision or vagueness of data. Both radial bases neural networks and fuzzy logic approach, though they represent different techniques, were shown by Kecman[12] to be mathematically very similar.

The real learning systems must satisfy the epistemological commitment in order to be 'intelligent' [15]. Current learning systems have actual limiting problems: the generalization problem or the problem of over learning, the problem of inductive bias in learning, and thirdly the empiricist dilemma or understanding constraint-free evolution [15].

The introduced learning models are usually constrained by their structure, this represents an approximation; how could we know that a given structure is suitable, not larger or smaller, of certain application? Another problem is the 'quality and quantity' of training data, for example how could we know that the training data are sufficient to fully represent the whole problem domain? [15]. Another problem is that further training can lead to the problem of over-fitting, or over-training, it is not trivial to know when to stop training in order to prevent this problem [15]. The problem of using neural networks in general is that we cannot prove that it reaches optimal approximation of required function every time we train it. The second problem, the inductive bias or the rationalist's a priori, reflects the bias of the creator of the learning model to see the world as he expects to be. For example, "In neural networks, learning behavior also assumes an inductive bias. For example, the limitation of perceptron led to the introduction of hidden nodes" [15]. The third problem, the empiricist's dilemma, discusses the problem of convergence of unsupervised learning models; we actually do not know where we are going. The unsupervised learning models also have the problem of inductive bias, which appears in the design of nodes, selection operators, and search techniques.

Learning in engineering terms can be defined as function approximation. In general, a function means dependency of some variables; the casual relation between variables is also known to be a function [27]. What makes the function approximation difficult is that there is no theoretical background of determining what the best form to approximate the actual function. In function approximation, the error function does not depend linearly on weights, hence the convex property is not always satisfied and then the convergence of these approximations is not always guaranteed [27]. The engineering problem of function approximation is that we have to approximate or interpolate a set of sparse and noisy training data points. Models used for functions approximation are known as networks or machines [27].

In function approximation, the two basic problems are choosing the form and the norm [27], by form we mean the function that can approximate the actual unknown function, by norm we mean the distance function that measures the goodness of the approximation function. The traditional functions used for functions approximation are tangent hyperbolic, radial basis function (Guassian, multi-quadratic), polynomial functions, fuzzy membership function (triangle, trapezoidal, singleton), bases, activation (multiplayer perceptron, RBF, regulation network), and truncated Fourier series [27]. The historical approximations are algebraic and trigonometric polynomial defined over the whole domain, but they limited capabilities of taking sharp bends followed by flat ones. Piecewise functions divide the region into several intervals by set of joints. When the positions of these joints are subject to learning, the learning process is complex and requires a nonlinear optimization. When approximation methods, such as polynomials and splines, have the same number of parameters, they give the same approximation. The problem is that their Vandernonde matrix, if not singular, is very ill-conditioned [27].

When using a linear-in-parameter approximation function, the resulted error function is convex, when L2 norm is used, which implies a guaranteed global minimum. As the order of the approximated function increases, a better approximation is introduced. However, higher degree can lead to the over-fitting problem. The degree of the approximated model can be represented by the degree of the polynomial, number of nodes in the hidden layer in neural networks, or the number of fuzzy rules in a fuzzy model. Over-fitting can be avoided by relaxing the interpolation requirements.

The problem of choosing a form is much more important than choosing a norm, since if the function of approximation is not compatible with the underlying function, then none of the norms can improve its bad approximation [27]. The choice of the norm depends on the practical application in hand. The most difficult problem in function approximation is determining the number of parameters or number of neurons in neural networks. The number of neurons in hidden layer determines the capacity of the networks and its ability to approximate a certain function. Furthermore, the error functions of soft models are not convex or quadratic, because the error function depends nonlinearly upon the weights, and hence the search for minimum is a hard and uncertain task.

To summarize, in function approximation there are always two extremes that should be avoided: Filtering the relations itself and hence eliminating some of its characteristics, and other extreme of following the noise and overfitting the training data. In our approach, learning is not a computational process that is defined over a predefined mathematical model, rather it is based on local interactions and local decomposition of signal effects as it will be described later.

## 4. Perception Logic

What we perceive in our daily life is not a Boolean approximation of the pair of opposites or extremes, but a continuous and concurrent perception of both extremes. In this analysis, the real number that represents a continuous quantity, in our world, can be perceived by a pair of '*observables'* or quantities; each represents the degree of belonging to either of the two extremes.

One of the main difference between what we call *perception logic* and other existing logics is the view of extremes, for example the view of white and black, good and bad etc. All existing logics express their logical values as 1 and 0, we argue that this is a part of the truth, the whole picture could be clear if we express for any object the logical value of it with respect to each extreme, in other words the white will take the perception value of -1 and the black will take the perception value of +1. Any perception value in the range [-1,+1] has a degree of belonging to each of these extremes that fully represent its position in the perception space.

It is known that humans do not have the self ability to measure anything accurately similar to any metric device, and they tend to give fuzzy words to express a measurable quantity, but this does not imply that the signals in the brain are based on fuzzy logic, rather, as we argue, the perception signal, inside the brain, fully represents the actual signal, but there is no association of it with an exact word, a symbol, or even a number. Therefore, fuzziness comes as a result of the inability to associate a symbolic number to any observed stimulus. However, in the planning phase of cognition, when a decision is required to be taken based on incomplete information or knowledge, we may expect that, in this phase, something similar to fuzzy inference engine exists.

Measuring can be viewed, in terms of perception or cognition terms, as a mapping from a natural quantity to a symbol; this mapping, from one quantity to a symbol, can be represented by dedicating a certain neuron, and hence it costs a delay in the perception propagation time. Since most of the shortened cognition cycles (only sensing, perception, and action steps) are performed in real time, we expect that there is no intermediate measuring step along this cycle, we know that most of our activities and responses are done without thinking (or doing the planning step), or observing (acquiring new sensing information, or shortly the acquisition step).

If you give a person a gray paper (not white or black) and ask him: Is it white? He may answer: *yes* by 70% (and consequently *black* by 30%) or he may answer "I *have no idea*". This answer does not mean that his answer is ambiguous or even fuzzy, instead, he gave an exact answer based on his knowledge of what white and black colors appear, or he did not make you confused when he actually does not know any thing about colors and says "I have no idea". It should, herein, be noted that switching in logic design is accomplished by assuming the existing of the *high impedance* state, or the *third* state. The same concept could be assumed in neural network design.

In order to illustrate our idea, let us give the following example: If we ask a hypothetical machine, which is modeled based on a certain logic, a question "is the weather hot?", the response of the machine will depends on the type of logic that is based upon, as illustrated in table (I):

| **The machine's answer** | **Type of logic** |
|---|---|
| *yes*/*no* | Boolean logic |
| *yes*/*unknown*/*no* | Three-value logic |
| *yes* (by level 6 out of 10 levels) | Multi-valued logic |
| *yes* by 70% and *Low* by 50% !! | Fuzzy logic |
| *yes* by 70% (which implies that it is cold by 30%) /*null* (ideal or passive) | Perception logic |

Table (4.1)

All these types of logic can be certainly simulated by electronic computers. It should be noted that basically Boolean logic does not contain the third state that tells us that the answer 'ideal' or 'no comments', however the actual digital computer architecture incorporates this state in the design of its circuits, which could motivates us to think that digital computers may have more computational power than Turing machine has.

The three valued logic has its ambiguity in defining its operations, multi-valued logic was argued by many authors as being a generalized version of Boolean logic, fuzzy logic, irrespective of its success, still has ambiguity, ambiguity implies that the machine did not have an exact measure, this can be a result of incomplete knowledge.

The perception logic, incorporate the use of the *truth* values of the two extremes at the same time, which are the complement of each other at any time (no ambiguity is assumed), the reason behind this is that they will be interpolated, as it will be shown later, in order to approximate any function or behavior!!, moreover it includes the *null* state as well. It should be noted that *ideal* or *passive* is much similar to the engineering use of the *third state* or *high impedance* in digital circuit design, For example, in regular expressions, we have λ which represents a string of length zero, and Ø which means no string at all, and in logic circuit design, the engineers were enforced to add a third state, or what is called 'high impedance' in order to solve the problem of switching such as sharing a bus, etc.

Therefore, it might be helpful to assume that, in the *architecture* of biological neurons, the neuron can be either excited with some level of excitation or not excited at all, which implies that it is not taking part in this perception process. In this logic, the *truth* and *falseness* are dual to each other, the duality implies the existence of truth and the falseness at the same time or concurrently. *What we interested in is how to manage this behavior in order to reach a point of learning in all what it means*,

## 4.1 A Foundation for Perception Logic

In this work, so far, it is not intended to build a theoretical foundation for this logic, rather, what we are interested in is to provide a necessary mathematical foundation that could be used as a base for developing automata that could model the perception process; that relate perceptual events and associate stimulus to responses in a cognitive structure like a brain.

Similar to the quantum bit, the perception *observable* can be viewed as an object that has both discrete and continuous aspects, like Qubit, this ***C-bit*** can be viewed as a quantum two-state system. The bases for this logic are the opposite value +1,-1 rather than the traditional orthogonal bases. A bit in the traditional Boolean logic has two separate states 0, 1, conversely the C-bit, or the perceptual observable, holds the two states +1,-1 simultaneously, any perception value is the superposition of these two states. Let $l_x$ and $l_{\neg x}$. $l_x$ represents the degree of belonging of the perceptual signal to the positive opposite +1, whereas $l_{\neg x}$ represents the degree of belonging to the negative opposite –1.

At any realistic perceptual observation:

$$l_x + l_{\neg x} = 1 \qquad (4.1)$$

For a linear belonging function $b_x(x)$, as shown in figure (5.1), the perception is a superposition of the opposite states +1, -1, i.e.:

$$x = l_{\neg x}(-1) + l_x(+1) \qquad (4.2)$$

or $x = (1 - l_x)(-1) + l_x(+1)$ (4.3)

and hence

$$x = 2l_x - 1 \qquad (4.4)$$

A C-bit, in this sense, carries the two possible bit value simultaneously. The C-bit projections of these opposites can be represented as:

$$cbit = \begin{bmatrix} l_x \\ l_{\neg x} \end{bmatrix} \qquad (4.5)$$

In general, the perception x, for any belonging function $l_x = b_x(x)$, we need a transformation T that transform $b_x(x)$ to a linear belonging function, the analysis of this transformation will not covered in this report, the perception observable of *Cbit*, i.e x can be formulated as:

$$x = \|cbit\| = \left\| \begin{bmatrix} l_x \\ l_{\neg x} \end{bmatrix} \right\| = T(l_x)(+1) + T_{\neg x}(l_{\neg x})(-1) \qquad (4.6)$$

Obviously, $T(l_x) = l_x$ for a linear belonging function. The complement of a *cbit* C can be defined as:

$$\overline{C} = \begin{bmatrix} l_{\neg x} \\ l_x \end{bmatrix} \quad (4.7)$$

And hence this relation holds:

$$\|\overline{C}\| = -\|C\| \quad (4.8)$$

Which means that the complement of the perceptual observable is *its negative value*.

As an example, the distinctive logical values for a white/black color can be given as (table(4.2)):

| **Color/logical values** | **x** | $l_x$ | $l_{\neg x}$. |
|---|---|---|---|
| *Completely black* | +1 | 1 | 0 |
| *Completely white* | -1 | 0 | 1 |
| *A gray color that does not tends to be white nor black* | 0 | ½ | ½ |
| *No color at all (transparent)* | Ø | Ø | Ø |

Table (4.2)

The OR operation between cbits of the same perceptual information can be given as follows:

$$C_1 \oplus C_2 = \begin{bmatrix} l_x^1 \\ l_{\neg x}^1 \end{bmatrix} \oplus \begin{bmatrix} l_x^2 \\ l_x^2 \end{bmatrix} = \begin{bmatrix} \frac{1}{2}(l_x^1 + l_x^2) \\ \frac{1}{2}(l_{\neg x}^1 + l_{\neg x}^2) \end{bmatrix} \quad (4.9)$$

It should be noted that the resulted Cbit preserve the property that the sum of its logical value is also 1 as indicated by equation (4.9).

Equation (4.9) directly implies:

$$\|C_1 \oplus C_2\| = \tfrac{1}{2}(\|C_1\| + \|C_2\|) \quad (4.10)$$

Which means that the OR operating between two *Cbits* can be don by averaging their perceptual information.

The AND operation between *Cbits* of the same perceptual information can be given as follows:

$$C_1 \otimes C_2 = \begin{bmatrix} l_x^1 \\ l_{\neg x}^1 \end{bmatrix} \otimes \begin{bmatrix} l_x^2 \\ l_{\neg x}^2 \end{bmatrix} = \begin{bmatrix} l_x^1 l_x^2 + l_{\neg x}^1 l_{\neg x}^2 \\ l_x^1 l_{\neg x}^2 + l_{\neg x}^1 l_x^2 \end{bmatrix} \quad (4.11)$$

Equation (4.11) directly implies

$$\|C_1 \otimes C_2\| = \|C_1\| \bullet \|C_2\| \quad (4.12)$$

The combination of two Cbits X & Y of deferent perceptual observables does not have only two bases but four bases: $+1_X+1_Y$, $+1_X-1_Y$, $-1_X+1_Y$, $-1_X-1_Y$. Which can be represented as their tensor product, i.e.:

$$\begin{bmatrix} l_x \\ l_{\neg x} \end{bmatrix} \otimes \begin{bmatrix} l_y \\ l_{\neg y} \end{bmatrix} = \begin{bmatrix} l_x l_y \\ l_x l_{\neg y} \\ l_{\neg x} l_y \\ l_{\neg x} l_{\neg y} \end{bmatrix} \qquad (4.13)$$

### 4.1.1 Logical functions

A logical function represents logical dependencies between an independent *Cbit*(*s*) and a dependent *Cbit*. The association (W) between a dependent observable (y) and an independent observable (x) can be given by:

$$y = f(x) = W \otimes x \qquad (4.14)$$

Which means that:

$$\begin{pmatrix} l_y \\ l_{\neg y} \end{pmatrix} = \begin{pmatrix} w_x & w_{\neg x} \\ \neg w_x & \neg w_{\neg x} \end{pmatrix} \otimes \begin{pmatrix} l_x \\ l_{\neg x} \end{pmatrix} = \begin{pmatrix} w_x l_x + w_{\neg x} l_{\neg x} \\ \neg w_x l_x + \neg w_{\neg x} l_{\neg x} \end{pmatrix} \qquad (4.15)$$

The elements of association vector $w_x$ and $w_{\neg x}$ belong to the range [0,1] can be determined as follows:

$$w_x = l_y \big|_{x=+1} \qquad (4.16)$$

$$w_{\neg x} = l_{\neg y} \big|_{x=-1} \qquad (4.17)$$

### 4.1.2 Logical spaces

The logical space of L(x), which belongs to the range [-1,+1], can be decomposed into two other logical space [-1,0], [0,1], namely L(x') and L(x''), the practical view of this analysis is when the observations of y y are given at x=-1,0,+1 respectively, i.e. at the boundaries and at the mid-point, the relation between the logical spaces as follows,:

$$L(x) = L(x') \oplus L(x'') \qquad (4.18)$$

Where x',x'' are relative observable and are related to x by:

$$x = \begin{cases} \frac{1}{2}(x' + 1) & x' \neq \emptyset \\ \frac{1}{2}(1 - x'') & x'' \neq \emptyset \end{cases} \qquad (4.19)$$

Or Inversely:

$$x' = \begin{cases} 2x-1 & x \in [0,+1] \\ \emptyset & x \in [-1,0] \end{cases} \quad \& \quad x'' = \begin{cases} \emptyset & \mathrm{x} \in [0,+1] \\ 2x+1 & \mathrm{x} \in [-1,0] \end{cases} \tag{4.20}$$

Let the operator $\lvert \mathrm{x} \rvert$ be defined as

$$|x| = \begin{cases} x & -1 \le x \le +1 \\ \emptyset & \text{otherwise} \end{cases} \tag{4.21}$$

Hence we can describe the logical variables can be related as:

$$x = \left|\tfrac{1}{2}(x' + 1)\right| \oplus \left|\tfrac{1}{2}(1 - x'')\right| \tag{4.22}$$

This analysis will be used to describe the relation between bases functions at different perception levels as it will be given in section (5.2.1).

### 4.1.3 Logical functions over logical subspaces

At the logical subspace *L*(*x*'), the logical relation between *y*' and *x*' can be given (as described before) as follows:

$$y' = W'x' \tag{4.23}$$

Similarly, at the logical subspace *L*(*x*''), the logical relation between *y*'' and *x*'' can be given as follows:

$$y'' = W''x'' \tag{4.24}$$

Since the L(x) spans the two subspaces L(x') and L(x''), we can say:

$$y = y' \oplus y'' \tag{4.25}$$

Or $y = W' \otimes x' \oplus W'' \otimes x''$ (4.26)

Or in the vector form:

$$y = \begin{pmatrix} w'_x & w'_{\neg x} \\ \neg w'_x & \neg w'_x \end{pmatrix} \otimes \begin{pmatrix} l_{x'} \\ l_{\neg x'} \end{pmatrix} \oplus \begin{pmatrix} w''_x & w''_{\neg x} \\ \neg w''_x & \neg w''_{\neg x} \end{pmatrix} \otimes \begin{pmatrix} l_{x''} \\ l_{\neg x''} \end{pmatrix} \tag{4.27}$$

And then:

$$y = \begin{pmatrix} w'_x l_{x'} + w'_{\neg x} l_{\neg x'} \\ \neg w'_x l_{\neg x'} + \neg w'_{\neg x} l_{x'} \end{pmatrix} \oplus \begin{pmatrix} w''_x l_{x''} + w''_{\neg x} l_{\neg x''} \\ \neg w''_x l_{\neg x''} + \neg w''_{\neg x} l_{x''} \end{pmatrix} \tag{4.28}$$

It should be noted that:

$$w'_{\neg x} = y'(x' = -1) = y(x = -1) = w_{\neg x} \tag{4.29}$$

And similarly:

$$w''_x = y''(x'' = +1) = y(x = +1) = w_x \tag{4.30}$$

And since the two logical space meet at point x=0:

$$w''_{\neg x} = y''(x'' = -1) = y(x = 0) = y'(x' = +1) = w'_x \tag{4.31}$$

$$y = \begin{pmatrix} w''_x l_{x''} + w'_{\neg x} l_{\neg x'} + w_{x=0}(l_{x'} \oplus l_{\neg x''}) \\ \neg w''_{\neg x} l_{x''} + \neg w'_x l_{\neg x'} + \neg w_{x=0}(l_{x'} \oplus l_{\neg x''}) \end{pmatrix} \tag{4.32}$$

Which means that:

$$y = \begin{pmatrix} w''_x & w_{x=0} & w'_{\neg x} \\ \neg w''_x & \neg w_{x=0} & \neg w'_{\neg x} \end{pmatrix} \begin{pmatrix} l_{x''} \\ (l_{x'} \oplus l_{x''}) \\ \neg l_{x'} \end{pmatrix} \tag{4.33}$$

Equation (4.33) states that the decomposition of the space L(x) into the two neighbors sub-spaces L(x') & L(x'') brings out three bases that can span the space as shown in figure (5.4). This result formulates the composition process of the subspaces L(x') & L(x'') in order to realize the function y = *f*(x) over the space L(x) based on the observations of y over x= +1, the mid-point 0, and -1.

This result constructs the basic milestone in our multi-resolution analysis of perceptual observables. The hierarchal decomposition of the logical spaces and the construction will be discussed later in section (5.1),

## 5. Perception Computing

Perception can be defined as the process of organizing information acquired from real world and construct a knowledge representation that enables the rational actions. This process includes the learning process as a main part. In order to model perception, we need a non-Turing machine model that is based on a different kind of logic, i.e the perception logic. Luger [15] has emphasized that: "It is possible to say: The universal machine of Turing and Post may be too general. Paradoxically, intelligence may require a less powerful computational mechanism with more focused control." [15].

Human beings or even animals do not learn based on assigning symbols to what they sense, or even giving a measure of it, actually they perceive without having the ability to measure sensing data. On the other hand, it is misleading to assume that the natural neural networks deal with numbers represented as symbols, as our digital computers or as Turing machine assumes. Our sensory systems receive various kinds of signals; they do not convert them into *symbolic numbers* as our daily computers do, rather a signal is converted into an information-observable (we borrow this term from quantum computing), this observable fully represents the original signal without any loss of information. After that, the neural network decompose the signal in a certain way that enables it to discriminate the signal at multi-levels of resolutions up to natural network capacity. The decomposed features are then associated with each others based on some learning mechanism; finally it is related to the desired action, which responds rationally to the outer world. Hence, neural networks can be seen as mechanisms for mirroring the outer world inside the brain and hence reacting to it. In spite of the outer world is not recast into an explicit symbols, we could expect that neurons are self-organized; this self-organization could represent the meaningful symbols or patterns that human being is used to recognize.

In our analysis, we assume that the sensory signal can vary between –1, which is actually –70 mv in nervous system, and +1, which is the max positive voltage out of a neuron, say +70mv. The value of +1 is the perceptual value of one extreme or opposite, like sweet, black, and soft, whereas the value of –1 is the is the perceptual value of the other extreme or opposite such as bitter, white, and hard. It is important to note here that the perceptual value zero corresponds to the neutral or the equilibrium state of a nervous signal that means neither sweet or bitter, hot or cold, black or white, and so on.

Therefore real quantities in real world space [-∞,+∞], for example *y*, is some how mapped onto the perception space [-1,+1], for example *x*, we can assume a function like *Tanh()* (equation(5.1)) perform this mapping: *y*→*x*.

$$\tanh(x) = \frac{e^{x} - e^{-x}}{e^{x} + e^{-x}} \qquad (5.1)$$

& its inverse $\tanh^{-1}$(x) is given by [20]:

$$\tanh^{-1}(x) = \frac{1}{2}\ln(\frac{1+x}{1-x}) \qquad (5.2)$$

In addition to these spaces, we also have a third space, the logical belonging space [0,1], in which each perceptual value is mapped separately onto two logical values $l_x$ and $l_{\neg x}$. $l_x$ represents the degree of belonging of the perceptual signal to the positive opposite +1, whereas $l_{\neg x}$ represents the degree of belonging to the negative opposite –1. It is clear that when $x = +1$, $l_x = 1$ and $l_{\neg x} = 0$, on the other hand when x = -1, $l_x = 0$ and $l_{\neg x} = 1$. If $x = 0$, then both , $l_x = l_{\neg x} = ½$ as shown in figure (5.1).

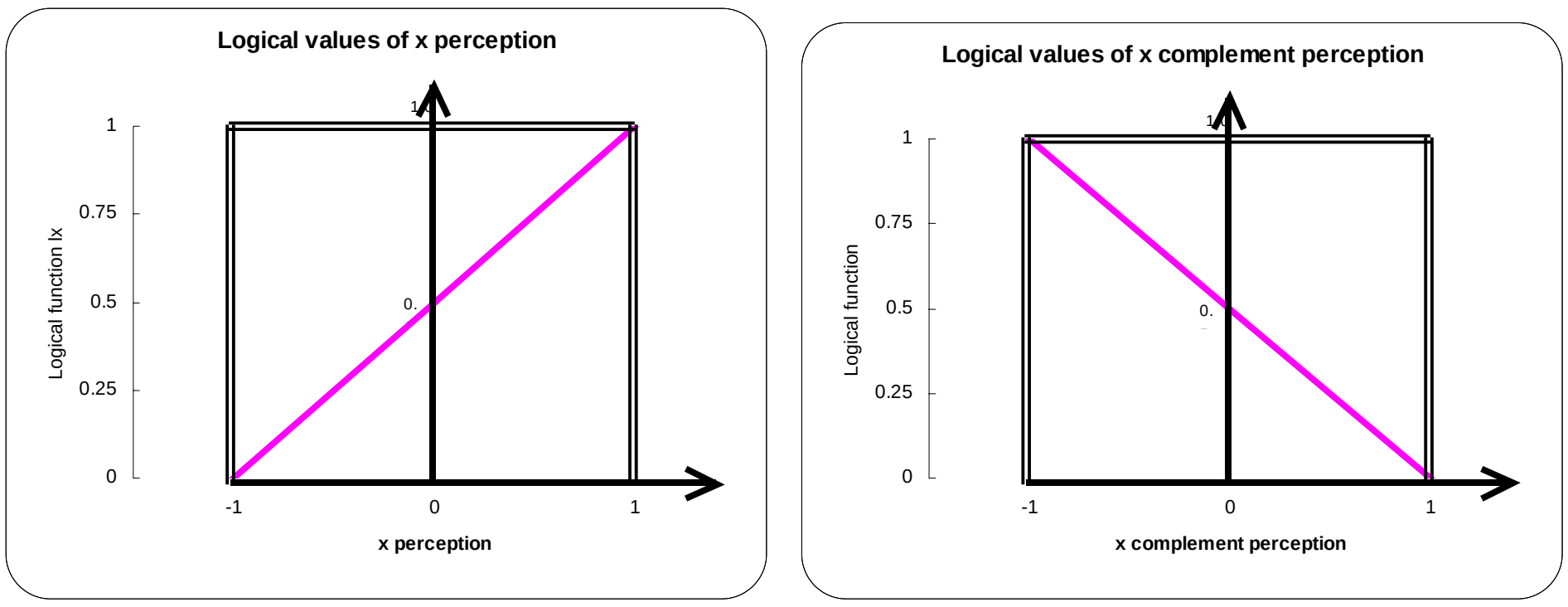

(a) (b)

Figure (5.1) : a) Logical values of $l_x$
b) Logical values of $l_{\neg x}$

Therefore we can say:

$$l_x = \tfrac{1}{2}(1 + x) \qquad (5.3)$$

And correspondingly,

$$l_{\neg x} = \tfrac{1}{2}(1 - x) \qquad (5.4)$$

Inversely, the perception function of a logical value x (figure (5.2(a)) can be represented as:

$$x = 2(l_x - \tfrac{1}{2}) \qquad (5.5)$$

And correspondingly (figure (5.2(b)):

$$\neg x = 2l_{\neg x} - 1 \qquad (5.6)$$

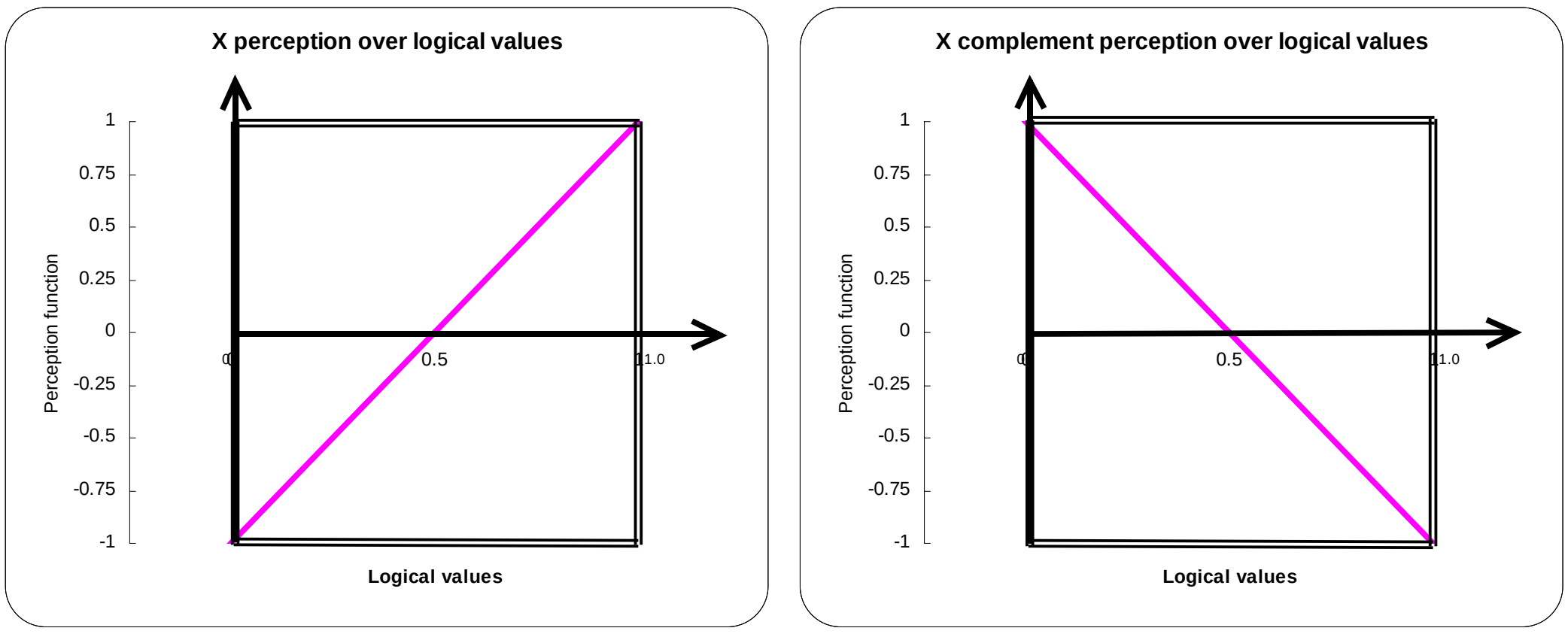

(a) (b)

Figure (5.2) : a) Perception x as a function of $l_x$
b) Perception ¬x as a function of $l_x$

## 5.1 Multi-resolution Analysis of Perceptual information

Similar to multi-resolution analysis in wavelets, the function is decomposed into several level of perception, each lover level gives a better image of this function until the full function image is perceived. In contrast, in this analysis, we have a father and a mother basis functions, namely $b_{+1}$ and $b_{-1}$ as shown in figure (5.3):

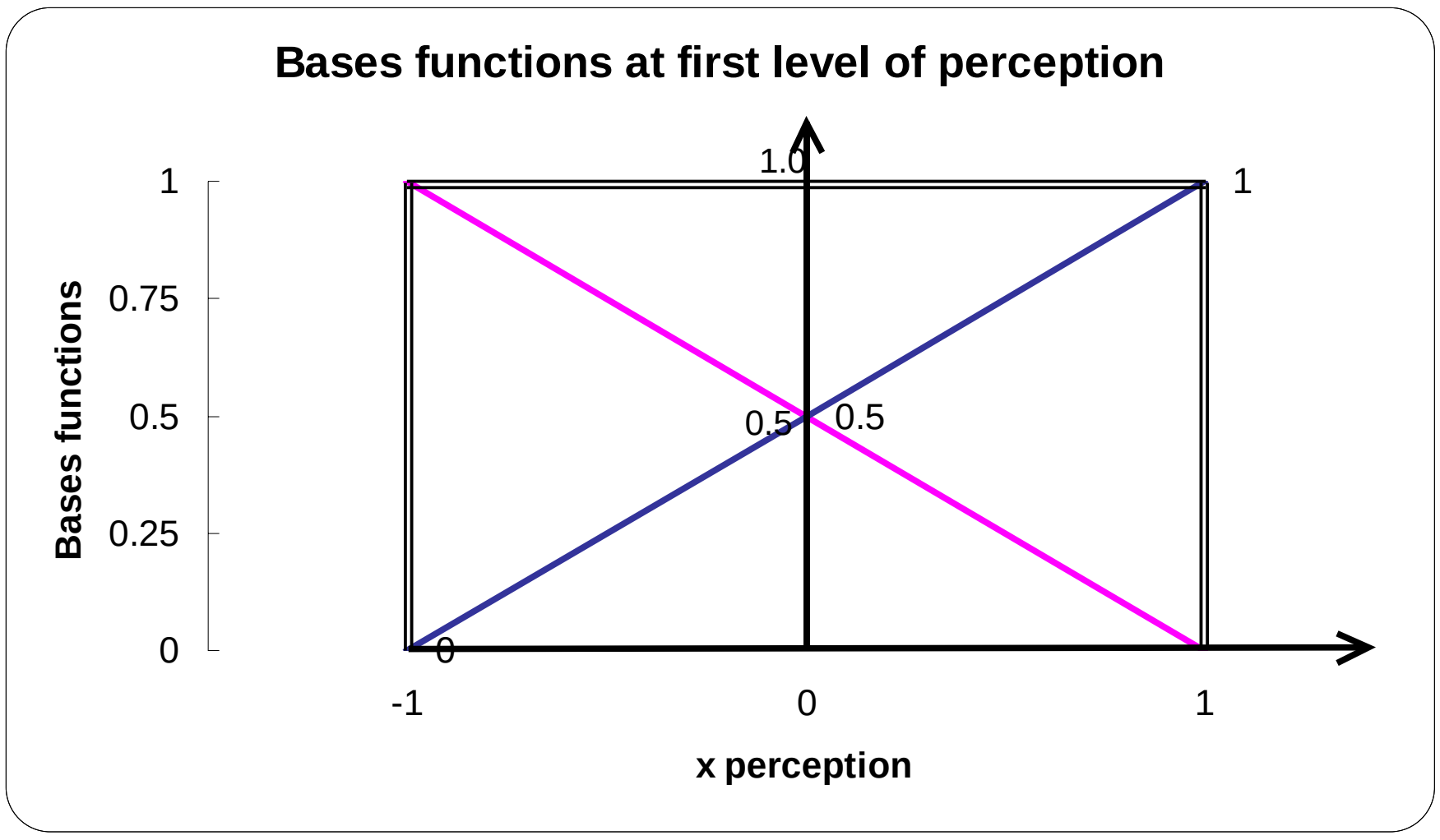

Figure (5.3) : Bases functions at the first level of perception

At the first level of perception, the function y=f(x) is approximated by its logical component $l_y$(as shown in equation 4.15), since $y$ and $l_{\neg y}$ can be completely reconstructed from it, as described in equation (5.7):

$$l_y = f^{(1)} = w_{-1}b_{-1}^1 + w_{+1}b_{+1}^1 \tag{5.7}$$

At the second level, as shown in figure (5.4), refer to equation (4.33), the second approximation of f is given by:

$$f^{(2)} = w_{-1}b_{-1}^2 + w_0 b_0^2 + w_{+1}b_{+1}^2 \tag{5.8}$$

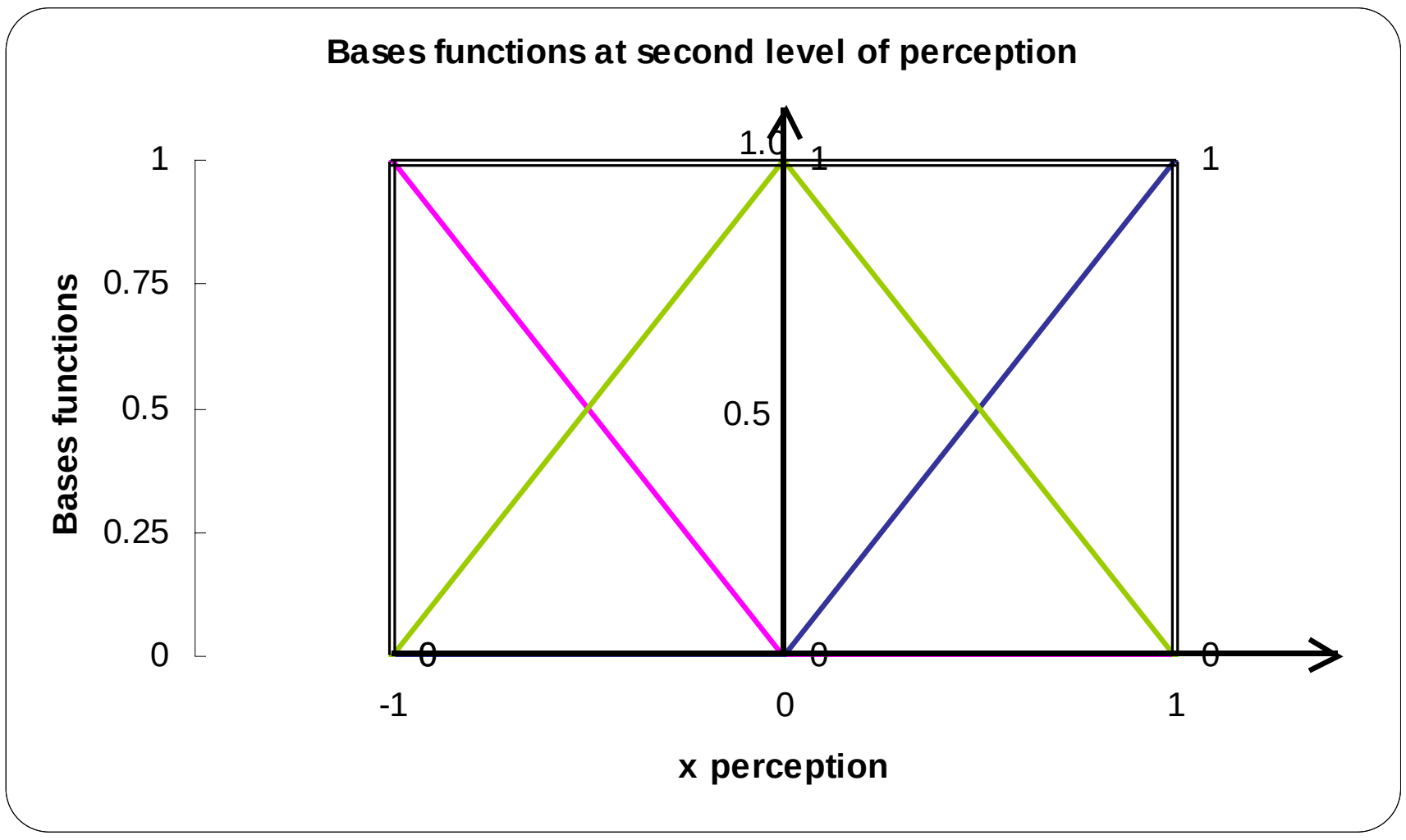

Figure (5.4) : Bases functions at the second level of perception

At this level we have 3 values $\{w_{-1}, w_0, w_{+1}\}$, at the second level they can be approximated by:

$$f^{(2)} = w_{-1}b_{-1}^2 + w_0 b_0^2 + w_{+1}b_{+1}^2 \qquad (5.9)$$

But at the first level of resolution, it is approximated as:

$$f^{(1)} = w_{-1}b_{-1}^1 + w_{+1}b_{+1}^1 \qquad (5.10)$$

And hence,

$$f^{(1)}(x=0) = \tfrac{1}{2}(w_{-1} + w_{+1}) \qquad (5.11)$$

At the same time:

$$f^{(2)}(x=0) = w_0 \qquad (5.12)$$

An important remark is to note that at x=+1, or -1:

$$f^{(2)}(x=+1, or -1) = f^{(1)}(x=+1, or -1) \qquad (5.13)$$

Therefore, the only difference occurs when x=0, and hence we can say:

$$f^{(2)}(x) = f^{(1)}(x) + (w_0 - \tfrac{1}{2}(w_{-1} + w_{+1}))b_0^2 \qquad (5.14)$$

We can check by substitution of values of x=-1,0,+1
Or, it can be expressed as:

$$f^{(2)} = w_{-1}b_{-1}^1 + w_{+1}b_{+1}^1 + w_0' b_0^2 \qquad (5.15)$$

Where the value of $w_0'$ is calculated directly as,

$$w_0' = w_0 - \tfrac{1}{2}(w_{-1} + w_{+1}) \qquad (5.16)$$

Since the values of $w_{-1}, w_0, w_{+1}$ lies in the range [0,1] then the maximum value of $w_0'$ {at $w_0=1$, $w_{-1}=w_{+1}=0$}=+1, and inversely the minimum value of $w_0'$ {at $w_0=0$, $w_{-1}=w_{+1}=+1$}=--1.

It should be noted that the value $w_0'$ represents the difference between the estimation of the function at level 1 and level 2. Therefore we can say:

$$f^{(2)}(x) = f^{(1)}(x) + d(f^{(2)}(x), f^{(1)}(x)) \qquad (5.17)$$

Similarly, at the third level of perception, as shown in figure (5.5):

$$f^{(3)}(x) = f^{(2)}(x) + d(f^{(3)}(x), f^{(2)}(x)) \qquad (5.18)$$

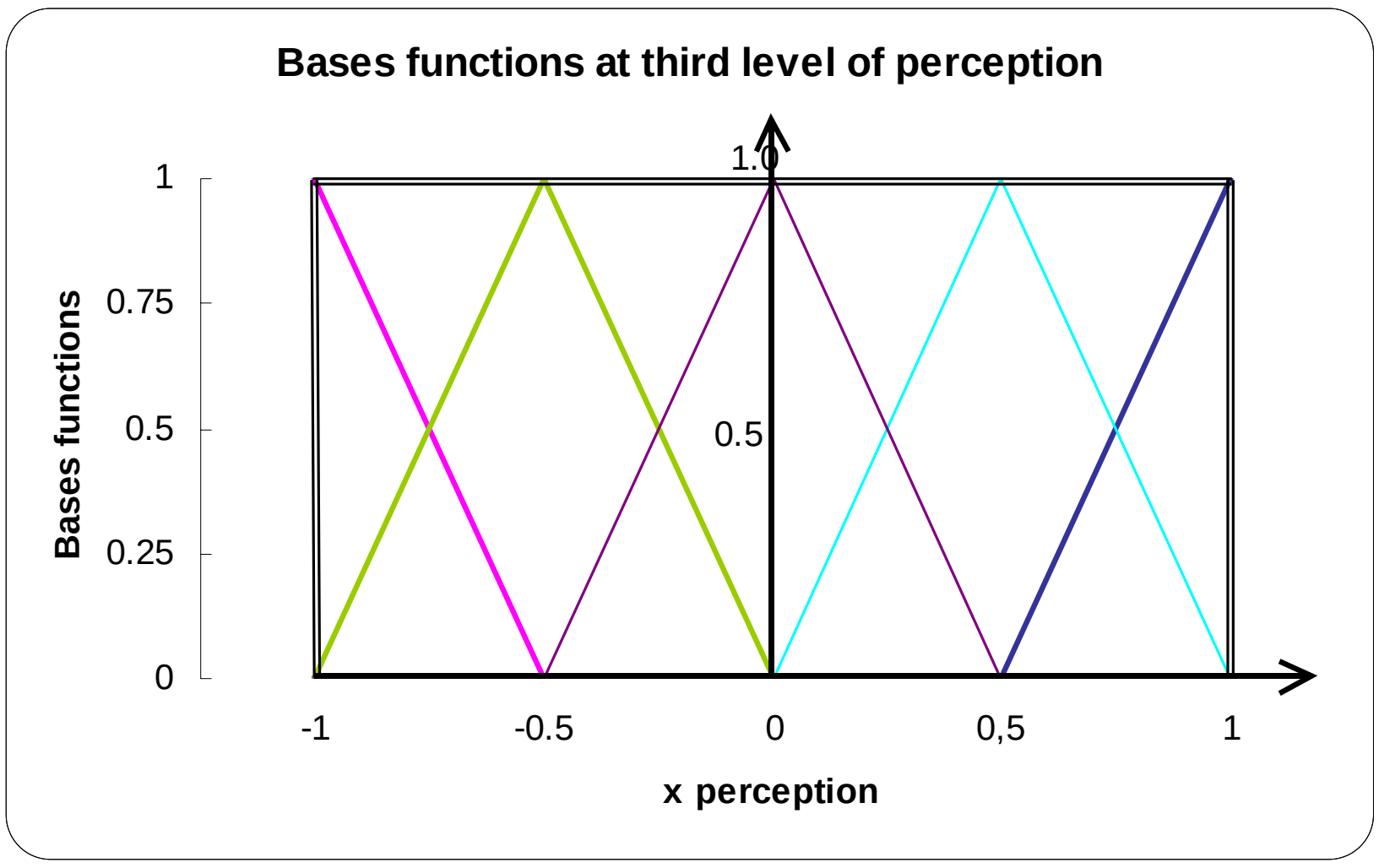

Figure (5.5) : Bases functions at the third level of perception

Where the difference between the approximation of the function at second and third level includes the addition of two basis functions at the mid of the new intervals, typically at -½ and +½ , specifically:

$$d(f^{(3)}(x), f^{(2)}(x)) = w'_{-\frac{1}{2}} b^{3}_{-\frac{1}{2}} + w'_{+\frac{1}{2}} b^{3}_{+\frac{1}{2}} \qquad (5.19)$$

The forth level well add at most 4 new basis at the middle of the intervals of the third level, specifically at -¾ , -¼ , +¼ , +¾.

And so forth. Therefore generally, at level *k*, we can say:

$$f^{(k)}(x) = f^{(k-1)}(x) + d(f^{(k)}(x), f^{(k-1)}(x)) \qquad (5.20)$$

Applying this equation recursively up to the first level yields:

$$f^{(k)}(x) = f^{1}(x) + \sum_{i=2}^{k} d(f^{i}(x), f^{i-1}(x)) \qquad (5.21)$$

The difference between functions at different perception level is given by:

$$d(f^{(2)}(x), f^{(1)}(x)) = w'_{0} b^{2}_{0} \qquad (5.22)$$

And secondly:

$$d(f^{(3)}(x), f^{(2)}(x)) = w'_{-\frac{1}{2}} b^{2}_{-\frac{1}{2}} + w'_{+\frac{1}{2}} b^{2}_{+\frac{1}{2}} \qquad (5.23)$$

And Generally, for i≥3:

$$d(f^{(i)}(x), f^{(i-1)}(x)) = \sum_{j=1}^{2^{i-3}} w'_{-\frac{2^{i-2}-2j+1}{2^{i-2}}} b^{i}_{-\frac{2^{i-2}-2j+1}{2^{i-2}}} + \sum_{j=1}^{2^{i-3}} w'_{+\frac{2^{i-2}-2j+1}{2^{i-2}}} b^{i}_{+\frac{2^{i-2}-2j+1}{2^{i-2}}} \quad (5.24)$$

Therefore these bases span the space as follows (table(5.1)):

| Perception space | -1 | -¾ | -½ | -¼ | 0 | +¼ | +½ | +¾ | +1 |
|---|---|---|---|---|---|---|---|---|---|
| Level 1 | $b_{-1}$ | | | | | | | | $b_{+1}$ |
| Level 2 | | | | | $b_0$ | | | | |
| Level 3 | | | $b_{-½}$ | | | | $b_{+½}$ | | |
| Level 4 | | | | | | | | | |
| Level 5 | | $b_{-¾}$ | | $b_{-¼}$ | | $b_{+¼}$ | | $b_{+¾}$ | |
| etc……. | | | | | | | | | |

Table (5.1) : Fundamental bases functions

The fundamental bases functions, till the third level of perception, are shown in figure (5.6):

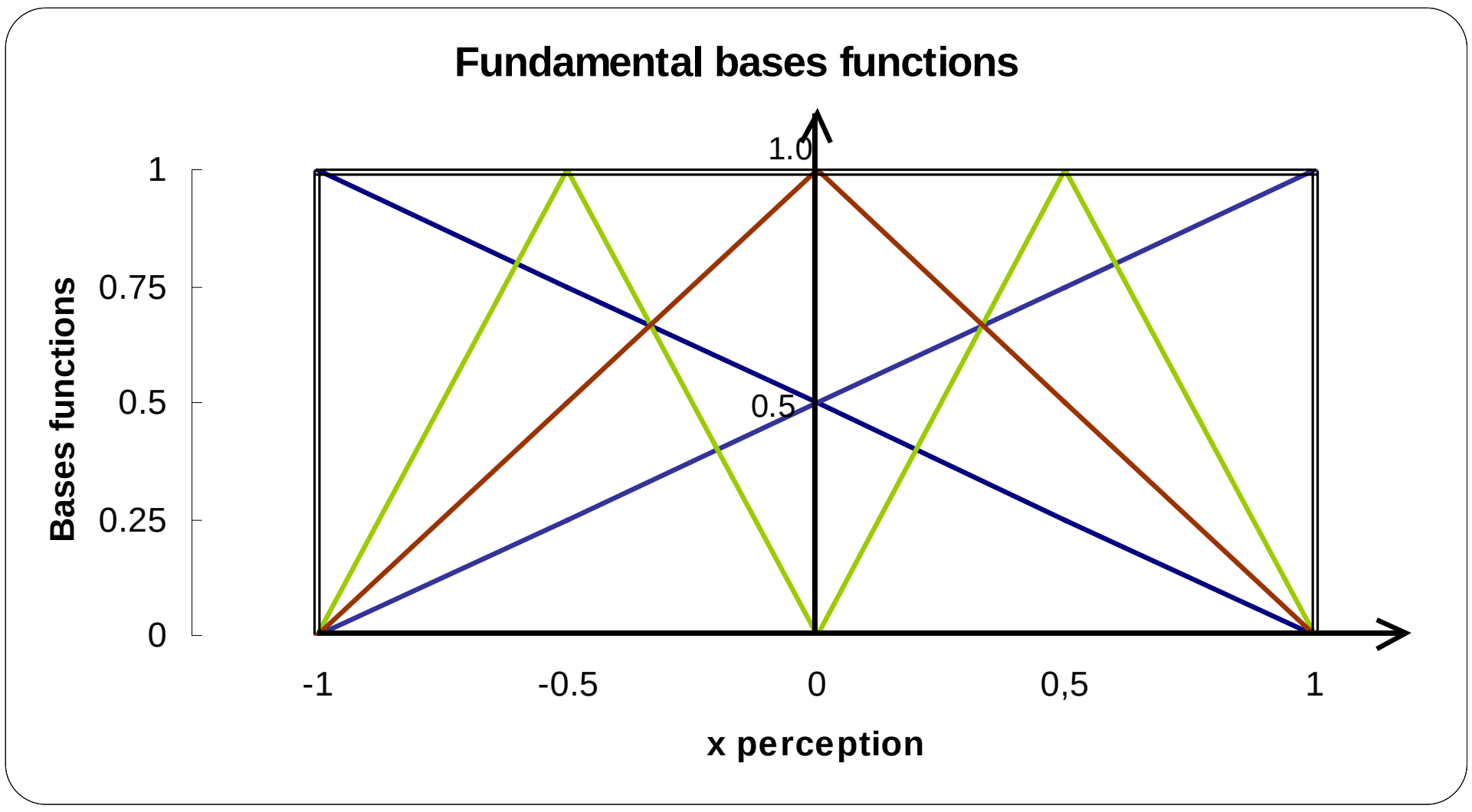

Figure (5.6) : Fundamental bases functions at first, second, and third levels of perception

## 5.2 Perception bases functions

The simplest base function could be:

$b(x) = l_x(x) = \frac{1}{2}(1+x)$ along with its conjugate $b_{\neg x}(x) = l_{\neg x}(x) = \frac{1}{2}(1-x)$

We can generally say:

$$b_x(x) = b_{\neg x}(-x) = 1 - b_{\neg x}(x) \quad (5.25)$$

Other example of a perception bases could be, as shown in figures (5.7), (5.8):

$$b_x(x) = \tfrac{1}{2}(1+\sin(\tfrac{\pi}{2}x)) \text{ defined only in the range [-1,+1].} \qquad (5.26)$$

The conditions over any perception bases could be:
a) $b(-1) = 0$
b) $b(+1) = 1$
c) $b(0) = \frac{1}{2}$
d) $b(x)$ is non-decreasing function over the range [0,1]

Further conditions could be added in order to ensure the smoothness of the bases functions such as:

e) $b'_x(x) = b'_{\neg x}(-x)$ at boundaries +1,-1. or more restricted conditions:

$$\frac{d^n b_x(x)}{dx^n} = \frac{d^n b_{\neg x}(-x)}{dx^n} \text{ at boundaries +1,-1}$$

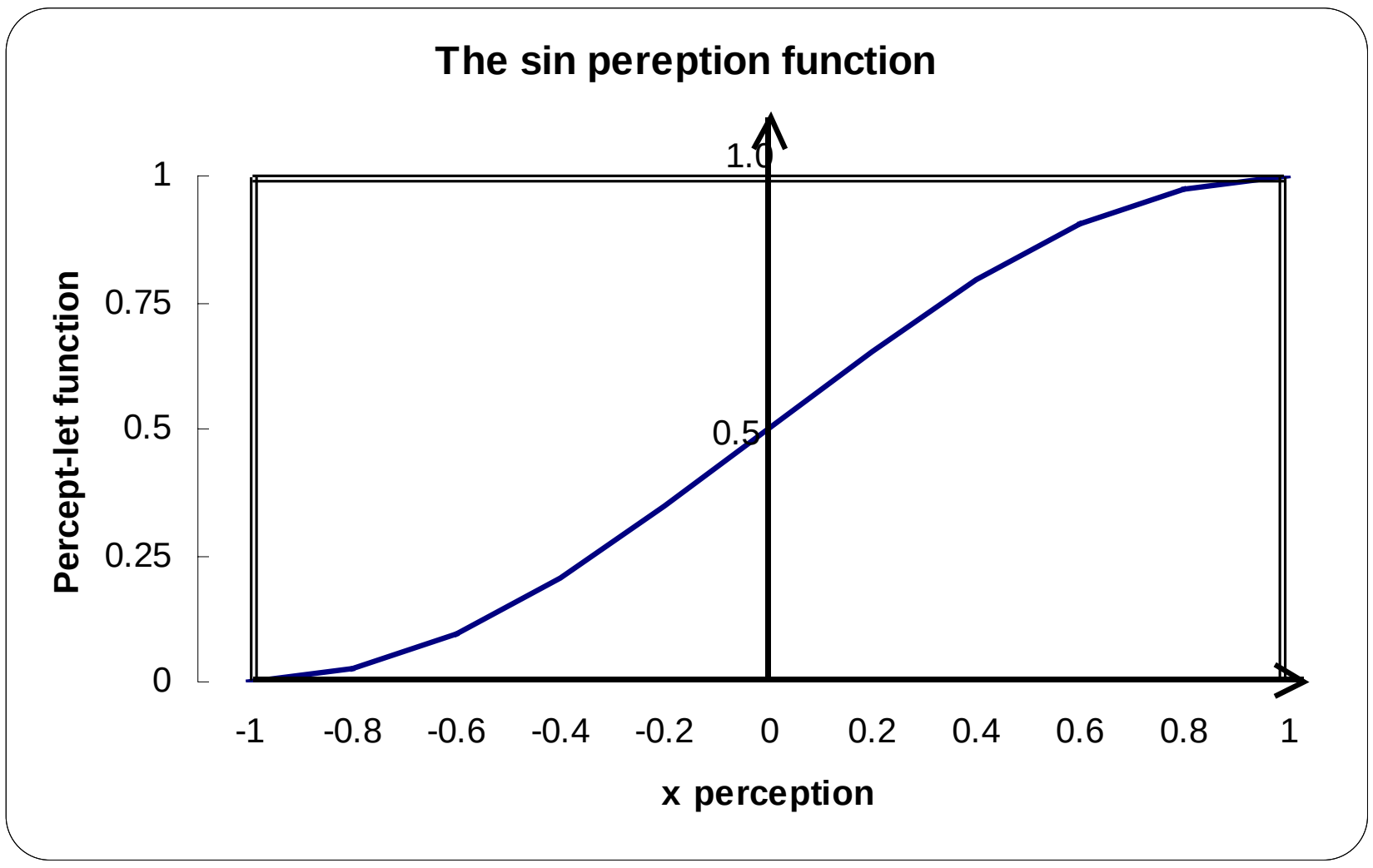

Figure (5.7): The sin *percept-let*

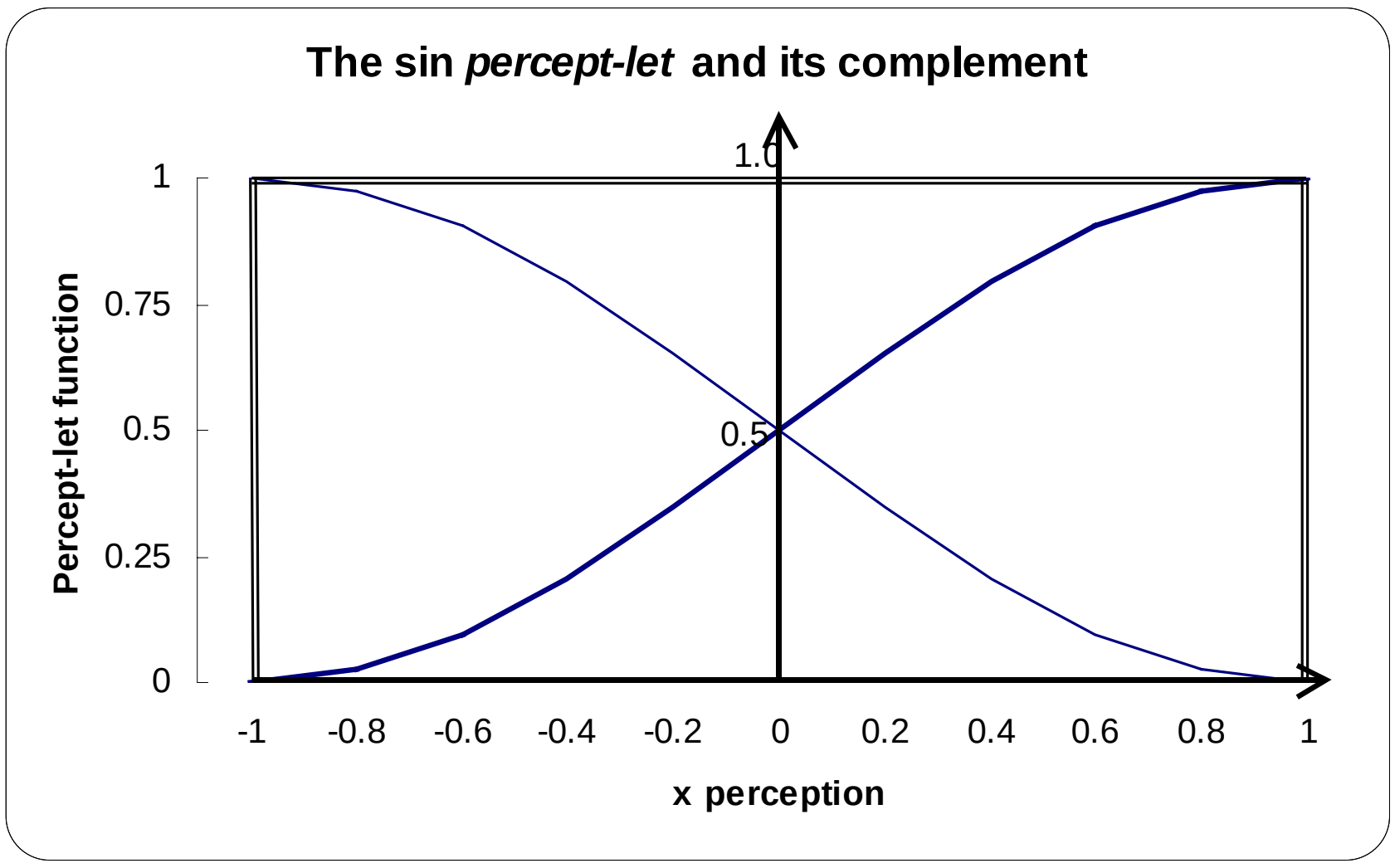

Figure (5.8): The sin percept-let and its complement

It should be noted that the sin *percept-let*, or shortly *perceptlet* satisfies all the above conditions.

### 5.2.1 ***Percept-let*** bases functions construction

Similar to wavelet analysis, the percept-lets can be formed recursively from one level to the higher level. In wavelets, this can be done by means of scaling and shifting; here we use these two operations, the difference of our analysis is that we have two parents functions in order to construct the bases as shown below and we need two types of shifting; shift to the right and shift to the left:

The bases of the first level of perception are the following:

The father percept-let:

$$b_{+1}^{1}(x) = b_{x}(x) \tag{5.27}$$

& The mother percept-let is its complement:

$$b_{-1}^{1}(x) = b_{\neg x}(x) = 1 - b_{x}(x) \tag{5.28}$$

Refer to logical spaces section (4.12), based on equation (4.20) describes the relation between the logical space and its sub-spaces, the bases of the second level of perception are the basic daughter percept-let and the other virtual (not fundamental) bases:

The basic daughter percept-let is given by:

$$b_{0}^{2}(x) = b_{+1}^{1}(2x+1) \oplus b_{-1}^{1}(2x-1) \tag{5.29}$$

The other virtual (not fundamental) bases are given by:

$$b_{+1}^{2}(x) = b_{+1}^{1}(2x-1) \tag{5.30}$$

$$b_{-1}^{2}(x) = b_{-1}^{1}(2x+1) \tag{5.31}$$

Alternatively, the mid-point base can be represented as:

$$b_{0}^{2}(x) = \neg(b_{+1}^{2}(x) \oplus b_{-1}^{2}(x)) \tag{5.32}$$

Similarly, the bases of the third level is given by:

$$b_{+1}^{3}(x) = b_{+1}^{2}(2x - \tfrac{1}{2}) \tag{5.33}$$

$$b^3_{-1}(x) = b^2_{-11}(2x + \tfrac{1}{2}) \quad (5.34)$$

$$b^3_0(x) = b^2_0(2x) \quad (5.35)$$

& The fundamental bases are given by:

$$b^3_{+\frac{1}{2}}(x) = \neg(b^2_{+1}(2x - \tfrac{1}{2}) \oplus \neg b^2_{+1}(2x - \tfrac{1}{2})) \quad (5.36)$$

& $b^3_{-\frac{1}{2}}(x) = \neg(b^2_{-1}(2x - \tfrac{1}{2}) \oplus \neg b^2_{-1}(2x - \tfrac{1}{2}))$ (5.35)

Or shortly,

$$b^3_{+\frac{1}{2}}(x) = b^2_0(2x - \tfrac{1}{2}) \quad (5.37)$$

& $b^3_{-\frac{1}{2}}(x) = b^2_0(2x + \tfrac{1}{2})$ (5.38)

Similarly, for the forth level:

$$b^3_{+\frac{1}{4}}(x) = b^3_0(2x - \tfrac{1}{4}) \quad (5.40)$$

$$b^3_{+\frac{3}{4}}(x) = b^3_0(2x - \tfrac{3}{4}) \quad (5.41)$$

$$b^3_{-\frac{1}{4}}(x) = b^3_0(2x + \tfrac{1}{4}) \quad (5.42)$$

$$b^3_{+\frac{3}{4}}(x) = b^3_0(2x + \tfrac{3}{4}) \quad (5.43)$$

And so forth, in general, at perception level i$\geq$ 3:

$b^i_{\frac{2^{i-2}-2j+1}{2^{i-2}}}(x) = b^i_0(x - \frac{2^{i-2}-2j+1}{2^{i-2}})$ for positive points and j ranges from 1 up to $2^{i-3}$

(5.44)

& $b^i_{-\frac{2^{i-2}-2j+1}{2^{i-2}}}(x) = b^i_0(x + \frac{2^{i-2}-2j+1}{2^{i-2}})$ for negative points. (5.45)

Or by relating the fundamental bases by the basic daughter percept-let:

$b^i_{\pm\frac{2^{i-2}-2j+1}{2^{i-2}}}(x) = b^2_0(2^{i-2}(x \mp \frac{2^{i-2}-2j+1}{2^{i-2}}))$ for j ranges from 1 up to $2^{i-3}$ (5.46)

### 5.3 Perception as a Transformation

Fourier transform as well as wavelets transform originally transform signals from time domain to frequency domain, the signal is a special type of function in which the independent variable is the time, the computations of these transformations depends on the assumption that samples are given at equal interval time, and this is not the case with our transformation. Other difference is that they require the existence of the samples at hand before the transformation begins whereas our transformation does not, therefore the perception transformation can learn on line and as the sample or data are provided. Additionally, our perception transformation can be viewed as a top-down approach for function decomposition, i.e. it starts by perceiving the whole linear term of the function, or the low frequency part (using the terms of signal processing), and then continues to realize the high order parts.

## 6. The Learning Process

Learning process includes, or models, knowledge acquisition about unknown system or concept, and it can be defined as an approach to estimate parameters of a model based on using training data, in engineering terms: it can be called parameter estimation, weights updating, identification, or self-tuning, this process "learning from experimental data" may be viewed as a statistical learning [27].

Learning can be viewed as the process of building a model that can describe certain phenomena, or seeking for a function that can fit all sample data, and it then proves to be efficient and general for the new coming data. This enables us to estimate, anticipate and predict the behavior of that phenomena. Learning or finding a suitable function can reduce a huge volume of sample data into a meaningful set of rules and relations, so it is like searching for general mapping function that coincides with training pairs values, and approximates the actual function for values in between or outside training data range.

Learning is not a trivial or simple task; it is still one of the most difficult parts in any A.I. model. However learning is considered one of the most important parts in any intelligent behavior [15]. Moreover any algorithm that can learn from data is said to be intelligent, the problem of learning is still addressed by many fields such as biology, neuroscience, psyclogy, and philosophy [27]. Up till now, the automation of knowledge acquisition using machine learning methods is still an active area of research [4],[16], the main objective of a learning model is to have models capable of thinking and acting rationally; a rational system should acquire knowledge and respond rationally at right times. Soft computing techniques that are able to learn, including fuzzy models, neural networks, and support vector machines, are recognized as alternatives to the standard hard computing approach [27].

The efficiency of the learning procedure depends upon its ability to abstract and summarize and to create a compact storage of the learnt functions, i.e. we do not have to remember a long list of if-then rules or to memorize a huge amount of pairs of data, the learning automata also depends also on its structure, which means that any new samples provided do not have to completely change or reinvent the structure [27]. Rather, it adds to it, without changing the structure or re-initializing the learning process.

We have two major cases of how data are represented:

a) Boundary perceptions: perceptions are given at the specific locations in which bases are located, i.e. data are located at centers of the bases, i.e. at -1, - ¾, -½, -¼,0, , +¼, +½ , +¾,+1 and so on.

b) Neighborhood perceptions: perceptions are not restricted to be in above centers.

The approximation of the function that relates perceptions depends directly on the required level of perception, or the *perception resolution*. The perception resolution could be defined as the maximum attained level of perception.

**6.1 Learning based on Boundary Perceptions:**

In this approach, the fastest learning process can occur if the samples are provided in the sequence that corresponds to the hierarchal structure of the perception levels, i.e. data at level 1 then data at level 1, and so on. Specifically, the sequence is given as follows {-1,+1},0,{-½,+½ }, {-¼,- ¾,+¼,+¾ }, … and so on.

Following this sequence, the number of epochs needed is typically *one epoch*,

But if the data is not given in the above order, we need to scan over the data to learn the bases of the first level, and so on till the maximum number of perception *pr*. Therefore, we need at most number of epochs of order O(log(N)) where N is the volume of input data, as shown in equation (7.2).

**6.2 Learning based on Neighborhood Perceptions:**

Assume that we have a set of sample perceptions in the form ($x_i$,$y_i$) for i =1 to *N*. In this case, we follow a Hebbian rule of learning in which we choose $w_x$ at base $b_x$ such that:

$$\underset{\mathrm{w}}{\mathrm{Min}} \sum_{i=1}^{N} (w_x . b(x_i) - y_i)^2 \qquad (6.1)$$

Each weight $w_x$ of a base $b_x$ at a level of perception *l* can be given by:

$$\hat{w}_x = \frac{\sum_{i=}^{N} b_x(x_i) y_i}{\sum_{i=}^{N} b_x(x_i)^2} \qquad (6.2)$$

In this case, the learning process is active starting from the first sample till the last one, so at sample number *n*, the weight of a given perception is given by:

$$\hat{w}_x(n) = \frac{\sum_{i=}^{n} b_x(x_i) y_i}{\sum_{i=}^{n} b_x(x_i)^2} \tag{6.3}$$

And at the next sample $n$+1:

$$\hat{w}_x(n+1) = \frac{\sum_{i=}^{n} b_x(x_i) y_i + b_x(x_{n+1}) y_{n+1}}{\sum_{i=}^{n} b_x(x_i)^2 + b_x(x_{n+1})^2} \tag{6.4}$$

$$\hat{w}_x(n+1) = \frac{1}{1 + \frac{b_x(x_{n+1})^2}{\sum_{i=1}^{n} b_x(x_i)^2}} \left(\hat{w}_x(n) + \frac{b_x(x_{n+1})}{\sum_{i=1}^{n} b_x(x_i)^2} y_{n+1}\right) \tag{6.5}$$

Let us define the learning indicator be defined as:

$$L(n) = \frac{1}{\sum_{i=1}^{n} b(x_i)^2} \tag{6.6}$$

Hence:

$$L(n+1) = \frac{1}{\sum_{i=1}^{n} b_x(x_i)^2 + b(x_{n+1})^2} \tag{6.7}$$

Or,

$$L(n+1) = \frac{L(n)}{1 + b_x(x_{n+1})^2 L(n)} \tag{6.8}$$

And the next estimation of $w_x(n)$ is given by:

$$\hat{w}_x(n+1) = \frac{1}{1 + L(n) b_x(x_{n+1})^2} \left(\hat{w}_x(n) + b_x(x_{n+1}) L(n) y_{n+1}\right) \tag{6.9}$$

After some algebraic simplifications:

$$\hat{w}_x(n+1) = \hat{w}_x(n) + \frac{b_x(x_{n+1}) L(n)}{1 + b_x(x_{n+1})^2 L(n)} \left(y_{n+1} - b_x(x_{n+1}) \hat{w}_x(n)\right) \tag{6.10}$$

Let $G(n+1) = \frac{b(x_{n+1}) L(n)}{1 + b(x_{n+1})^2 L(n)} = b(x_{n+1}) L(n+1)$ (6.11)

Therefore, the weight can be updated recursively as:

$$\hat{w}_x(n+1) = \hat{w}_x(n) + \Delta\hat{w}_x(n+1) \quad (6.12)$$

$$\Delta\hat{w}_x(n+1) = G(n+1)(y_{n+1} - b_x(x_{n+1})\hat{w}_x(n+1)) \quad (6.13)$$

Where $G(n+1) = b_x(x_{n+1})L(n+1)$ & (6.14)

$$L(n+1) = \frac{L(n)}{1 + b_x(x_{n+1})^2 L(n)} \quad (6.15)$$

And initially,

$$w(1) = \frac{b_x(x_1)y_1}{b_x(x_1)^2} = \frac{y_1}{b_x(x_1)} \;\&\; L(1) = \frac{1}{b_x(x_1)^2} \quad (6.15)$$

**6.3 Convergence of the Learning Data**

This learning paradigm is not heuristic, rather it always converge to the required function mapping.

Theorem 6.3

A function mapping *f*: x →y, given at points {–1,-($2^{pr-2}$-j)/($2^{pr-2}$),0,+($2^{pr-2}$-j)/($2^{pr-2}$) ,+1, for j=1,2,.., $2^{pr-2}$-1 } , can be fully represented by multi-perceptual representation given by equation (5.21).

Proof:

At the first level of perception, a mapping represented by the two points at the boundaries can be simply represented by the two bases at the boundaries multiplied by weights equals y's values at the boundaries. In general at level *pr* of resolution, it can be represented by the bases at this level, or $b_{-1}^{pr}$,.,..,$b_{+1}^{pr}$ with weights equal to the samples values given, and since this representation can be replaced by the bases of the above perception resolution *pr*-1 as shown in equation (5.20), consequently, the representation at level pr-1 can be replaced by the bases of the above perception resolution *pr*-2, and so on till level 1 as shown in equation (5.21). we can conclude that this function decomposition or transformation can represent any function.

We can view this function transformation as follows: The first level of perception extracts the linear part of the function, the second level of perception extracts the deviation of this approximation from the mid-point (at zero) and provides better approximation as if it adds a first term of non-linearity to the first resolution. Consequently, the higher level of perception compensates any deviation from the previous approximation until the function is fully represented by this decomposition process. In short, at the last level of resolution, the output is fully representing the function whereas at a lower resolution, the output approximates the function.

# 7. The Perception Automata

We use the term automata here to emphasize that the learning process of the proposed model is achieved by its motive power within itself. Obviously, this is an important feature that any intelligent system should have.

In this automata, the learning process is achieved locally within each node or neuron, the expansion of its structure, or the organization of its nodes ore neurons, is autonomous and does not need external or centralized guide.

Figure (7.1) shows the layout of the proposed automata, the inputs to this automaton are both $x_i$ and f($x_i$) for i=1,..,m (where m is the number of samples), as shown there are three phases, the decomposition phase, the learning phase, and the realization phase. The decomposition phase generates the bases of $x_i$. The learning phase updates the weights of the bases of each level in ascending phase, i.e. the set of weights of the first level $W^{(1)}$, namely { $w_{+1}$, $w_{+1}$} is updated first, then the difference of the sample value of the function and its resolution $\Delta f^{(1)}(x)$ after the first level is calculated, this difference is the input to the second level of learning. Consequently, at the second level of resolution, the set of weights $W^{(2)}$ { $w_0$} is updated, and the difference $\Delta f^{(2)}(x)$ is also generated, and then this process is repeated until the last level of resolution in which the difference $\Delta f^{(n)}(x)$ is expected to be negligible ($\approx$ 0).

At the last phase, the realization phase, $f^{(1)}(x)$ is added to the second estimate of the difference between, f(x) and $f^{(1)}(x)$, $f^{`(2)}$(x), and then this process is continued until $f^{(n)}(x)$ is generated.

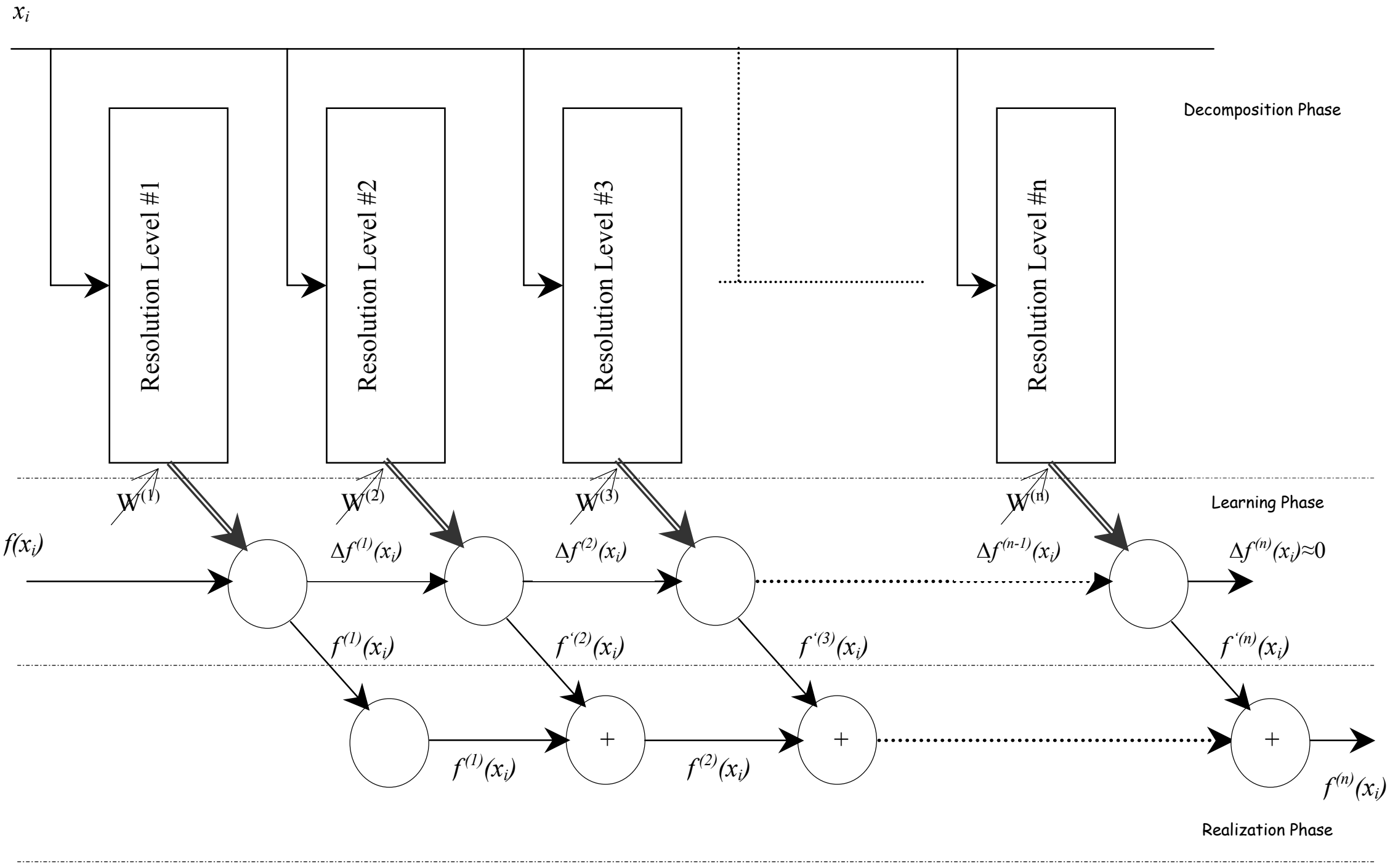

Figure (7.1) : The perception automata at learning process

At the realization step, as shown figure (7.2), no input f(x) is supplied since the function is realized up to level n of resolution, the objective is to give an estimate of it, the input $x$ is propagated over the resolution level and the estimate of the function at each is generated at each resolution by adding the previous estimate to the estimate of the difference at the current level till the function is fully estimated at the last level of resolution $f^{(n)}(x)$.

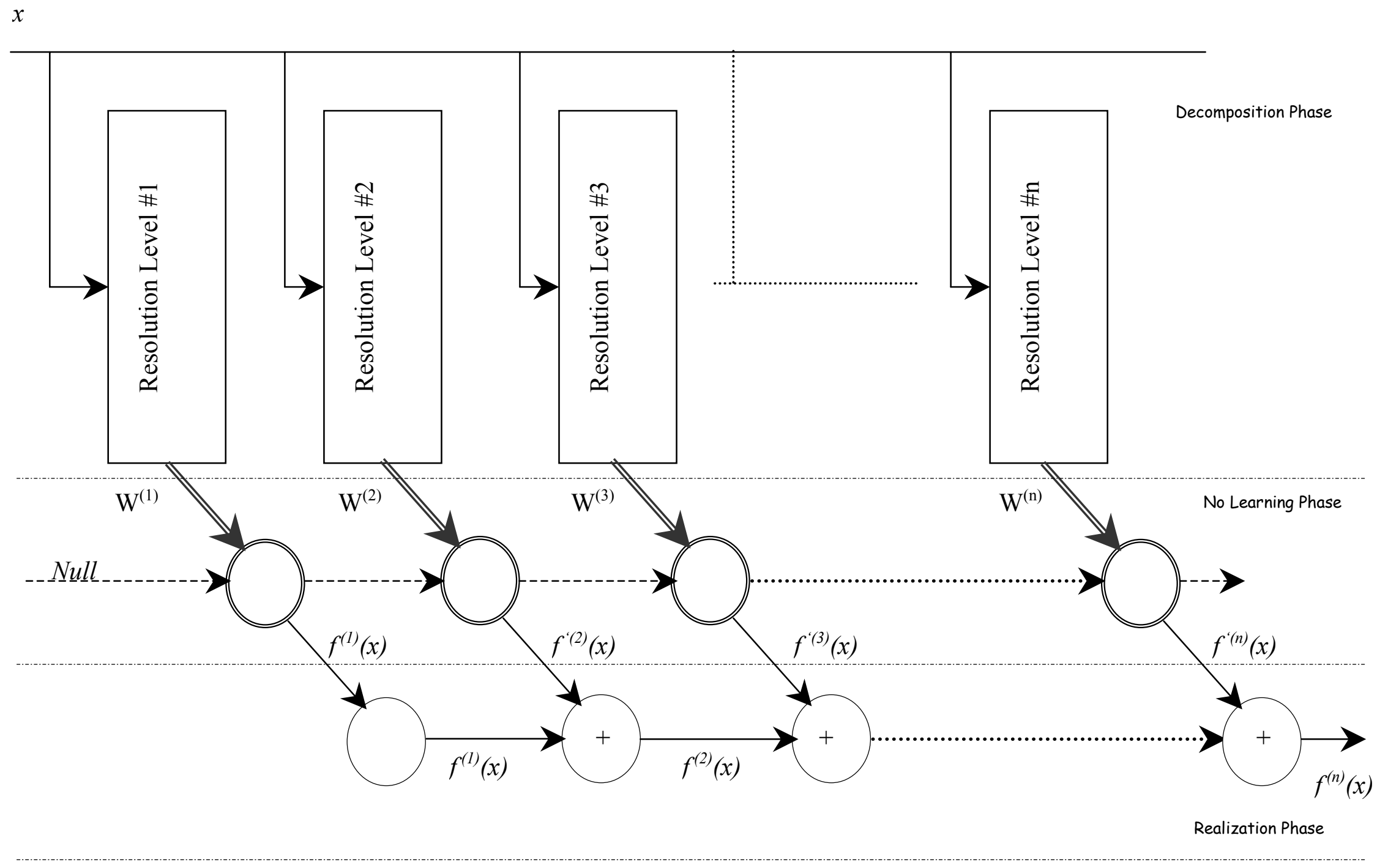

Figure (7.2): The perception automata at perception process

Figure (7.3) represent the learning step in more details, it shows the nodes of the decomposition phase at each level of resolution, for example, at the first level the transfer function is $b_{+1}(x)$ and $b_{-1}(x)$ respectively and their *synapses* $w_{+1}$ and $w_{-1}$. Similarly the details of the second and third levels of perception are presented.

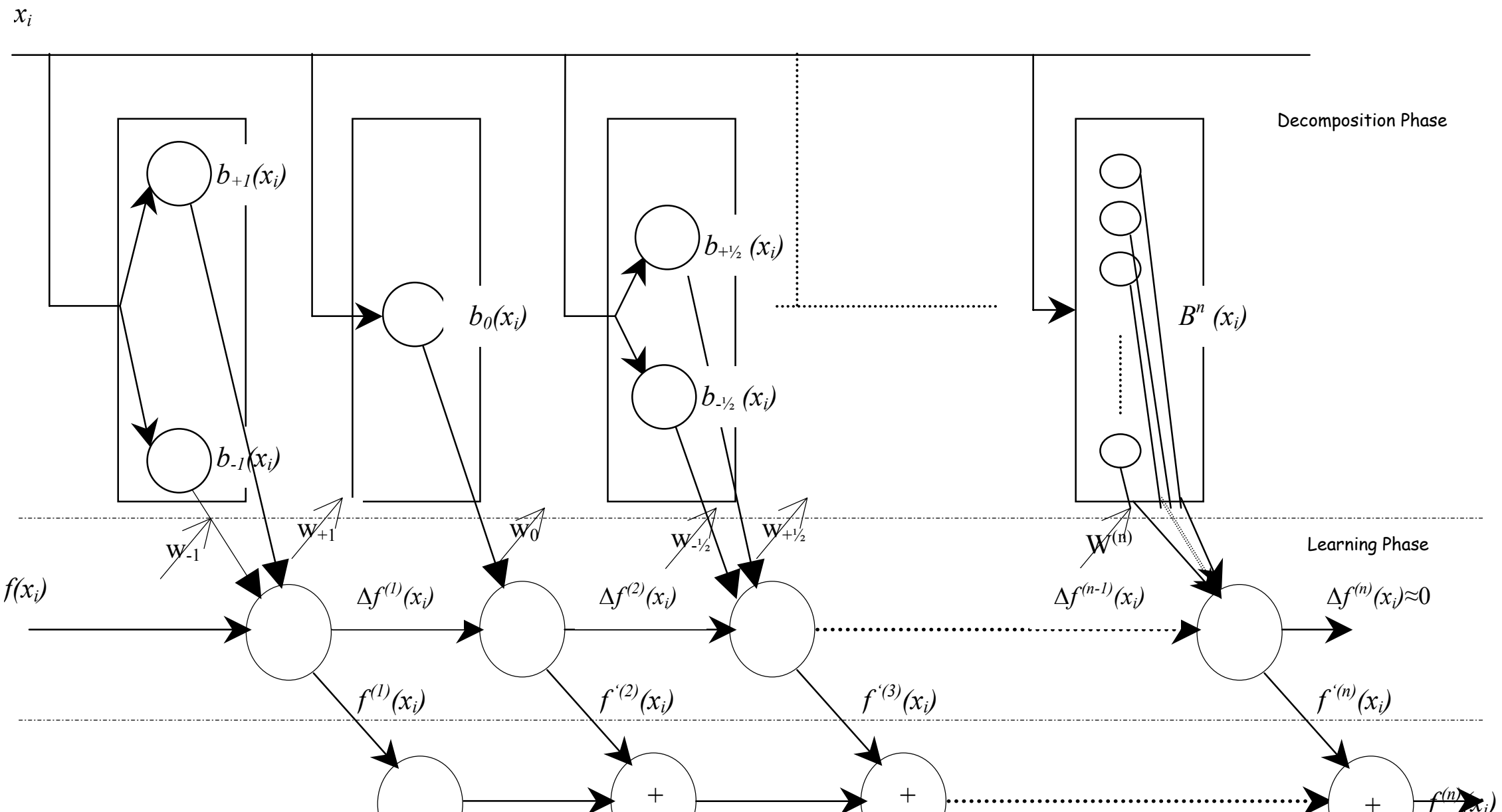

Figure (7.3): Learning step (detailed figure)

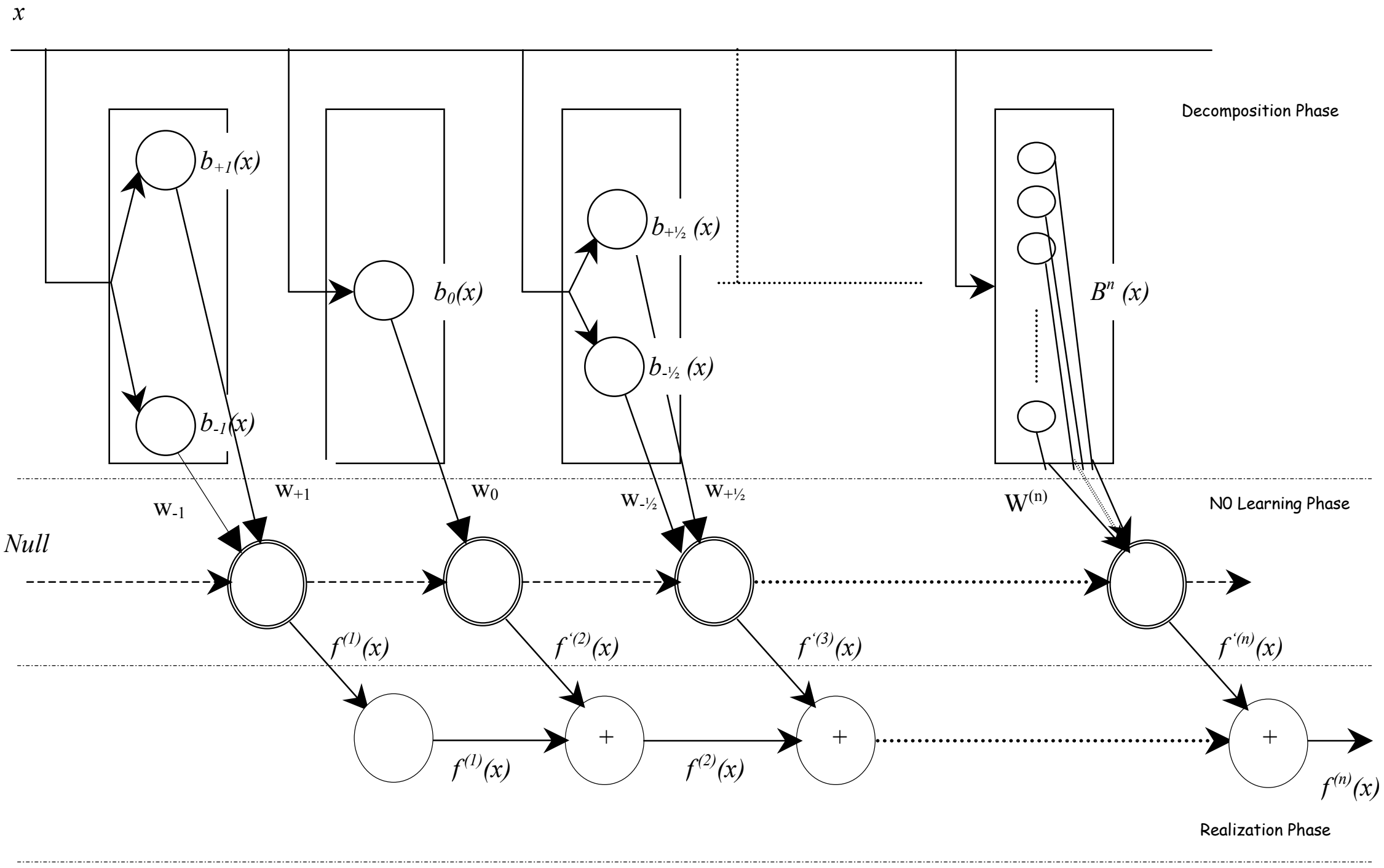

Figure (7.4): Realization Phase (detailed figure)

Figure (7.4) represents the realization step in more details, it also shows the nodes of the decomposition phase at each level of resolution, at this step no update is occurred to any weight, the learning step has already set final value to each weight.

### 7.1 Cognitive and Timing Complexity of the Perception Automata

Cognitive and timing complexity of this automata can not be studied in terms of size of input, the requirements and assumptions of this automata could be better described by the assumptions and requirements of *natural computing* as described by MacLennan [3]: The first requirement of any natural computing is real-time response, animals respond in a fraction of a second to an outer stimulus, this takes place in real-time, and therefore the speed of the basic operation is a critical issue. Intermediate results are not at all important; a progressive approximation that gets closer to the answer is more useful especially when it could anticipate a response before starting [3]. Analysis of computational model in terms of space and time complexity is less relevant in the context of natural computation since the size of input is generally fixed [3]. The comparison of natural computation models should be based on the criteria of the speed of response and generality of response (increasing the range of input yield better repsponce) [3]. Another criteria is the "*flexibility in response to novelty*", which measures the ability to respond correctly to a novel input [3]. Another criteria is "*adaptability*", since the natural environment is unpredictable and time-changing, the important issue is whether or not a natural system can adapt to a changing environment and how fast it can accomplish it. Gradual adaptation is generally preferred than abrupt change [3]. A further criterion is "*the stability of their learning*", Another criteria is "Tolerance to noise, error, faults, and damage", since natural environment is noisy, natural devices suffer from internal disturbance as well. [3].

Natural computation model has certain assumptions that must be fulfilled: first it should be physically realizable, i.e. its use of matter and energy should be finite, second assumption is that "the computation is governed by the physical aspects of representations", which implies syntactic formality [3]. The natural computing should be reduced to mechanical process, and intelligence should be defined in terms of specific process in order to deny the "ghost in the machine" assumption [3]. An important assumption is real-time response. Another assumption is the 'abstract formality', that is the dependence on the abstract forms of representations and their formal relationships. Further assumption is that all input, output, and information processing are all assumed to be continuous. Another assumption is the non-terminating property of natural computation; natural computation is in continuous interaction with its environment. It is better to think of natural computing as real-time control system, rather than computing a function such as in case of TM computation.An important assumption is the continual presence of noise, uncertainty, error, and indeterminacy in information representation and processing [3]. Final assumption is that input and output are assumed to be of fixed size. This is opposed to the unbounded representation of TM computation [3].

Regarding the cognitive complexity: if the perception resolution is $pr = 1$, then the maximum number of bases is only at –1, +1, which is 2, and at level 2 of perceptions, there is only one base at perception 0 and the number of perception $N$ is 3, therefore:

a) For $pr = 1$

Number of perceptions = 2

b) For $pr \geq 2$

= 2+1 if $pr = 2$
= 2+1+2 for $pr = 3$
= 2+1+2+4 for $pr = 4$
= $2+1+2+4+\ldots+2^{pr-2}$ for a perception resolution $pr$

or $N = 2 + \sum_{i=0}^{pr-2} 2^i$ for $pr \geq 2$

or $N = 2^{pr-1} + 1$ for $pr \geq 2$

Therefore, for all values of $pr$:

$$N = 2^{pr-1} + 1 \quad \text{for all value of } pr \geq 1 \qquad (7.1)$$

The perception resolution can be given in terms of N as follows:

$$pr = 1 + \log_2(N-1) \qquad (7.2)$$

Regarding the real-time perception (timing complexity) is simply $pr*d$ , where $d$ is the processing delay at a perception node.

## 8. Simulation Results

Assume that the function, which maps a perception x to the logical value of another perception f(x), is given by the disc-shaped dots in figures (8.1 ,8.2, 8.3). The samples are given at boundaries, i.e. at x values of –1, -7/8, -6/8,..,0,..,+7/8,+1, in this simulation, we used the *sin* percept-let. The function at figure (8.1) is nearly linear, the function at figure (8.2) has some nonlinearity, function, at figure (8.3), is more nonlinear. For all figures, the perception at each resolution level (from level 1 to level 5) is shown in each figure. Each function is not approximated, rather, exactly represented by the output of the maximum resolution level (in this simulation, the fifth one).

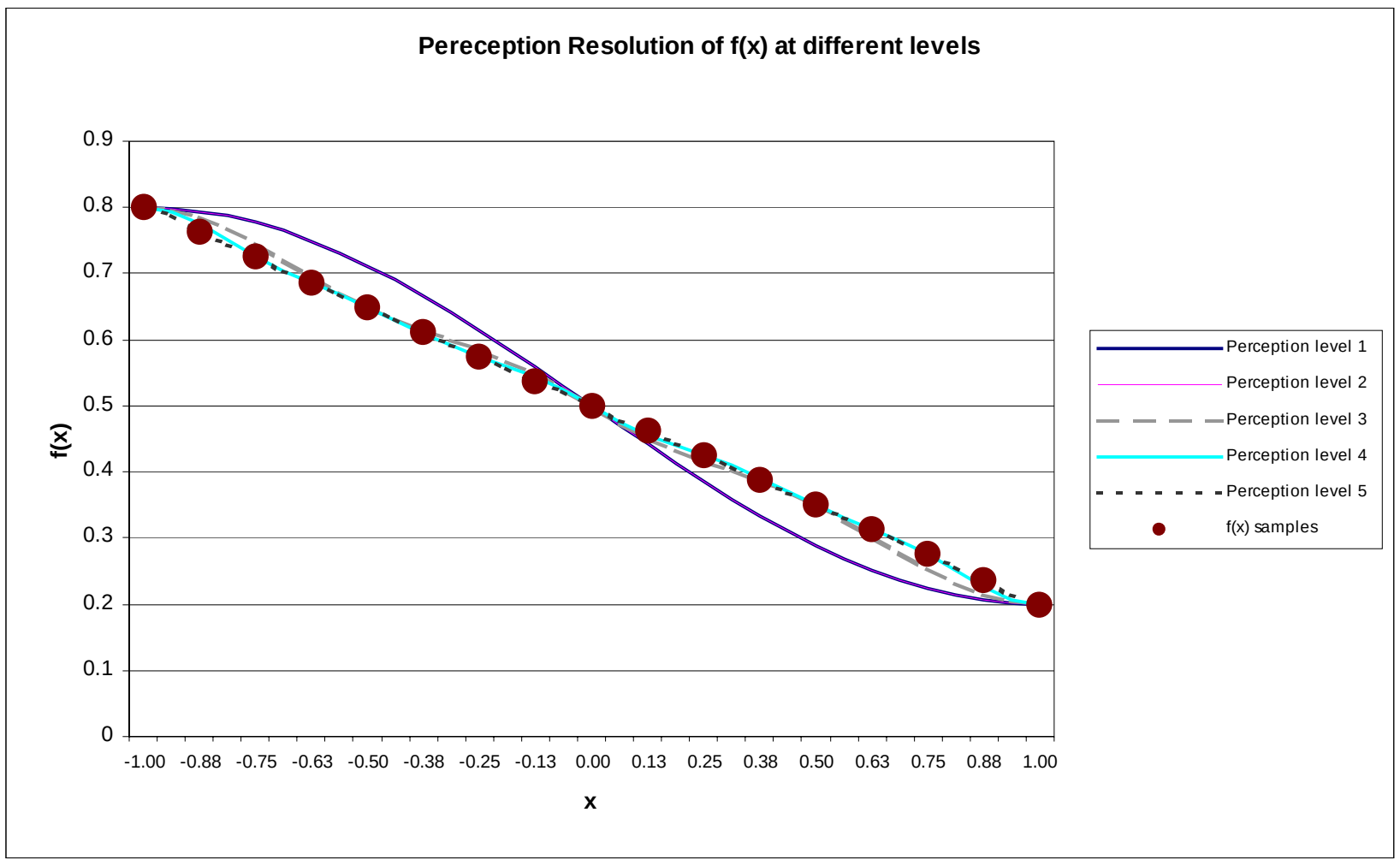

Figure (8.1) : A nearly linear samples approximated and modeled using multi-resolution perception

The number of epochs needed to 'teach' the automata is only 1 if the samples are sorted as described before, or at most 5 (maximum resolution level) if samples are not sorted.

Results of the third function were compared with feedforward neural network, a feedforward neural network was simulated using Matlab, using Levenberg-Marquardt (LM) optimization method for updating weights, and with 8 neuron having the *Tansig*() activation function. The performance of the network depends on the initial setting of weights, and on the average it reached a mean squared error (MSE) of 0.005 after 520 epochs (for the best run).

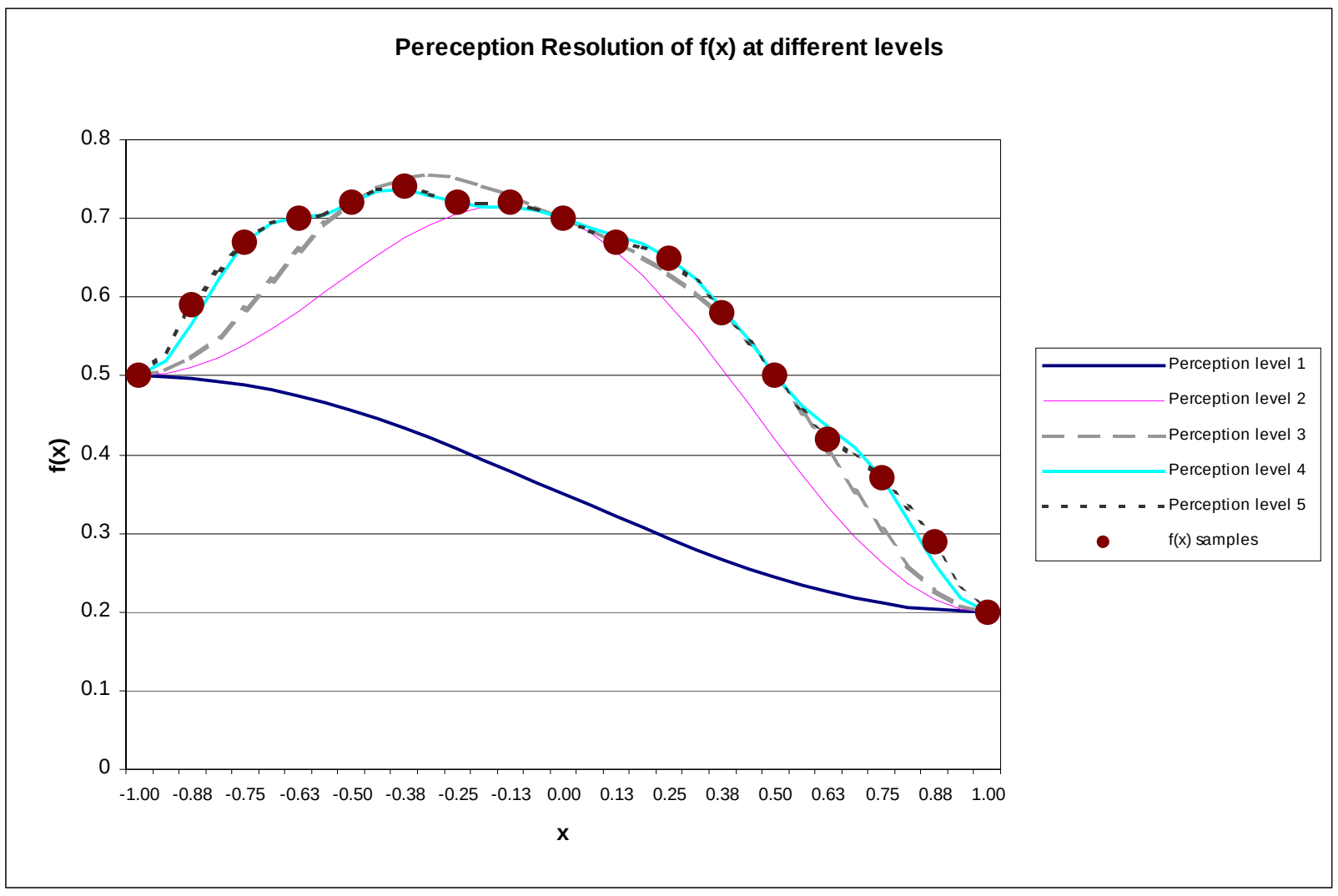

Figure (8.2) : A nearly quadratic nonlinear samples approximated and modeled using multi-resolution perception

By neglecting the weights of the fourth and the fifth resolution level the function of figure (8.1) can be mathematically represented as:

$$f^3(x) = 0.8b_{-1}^1 + 0.2b_{+1}^1 + 0b_0^2 - 0.062b_{-\frac{1}{2}}^3 + 0.062b_{+\frac{1}{2}}^3 \qquad (8.1)$$

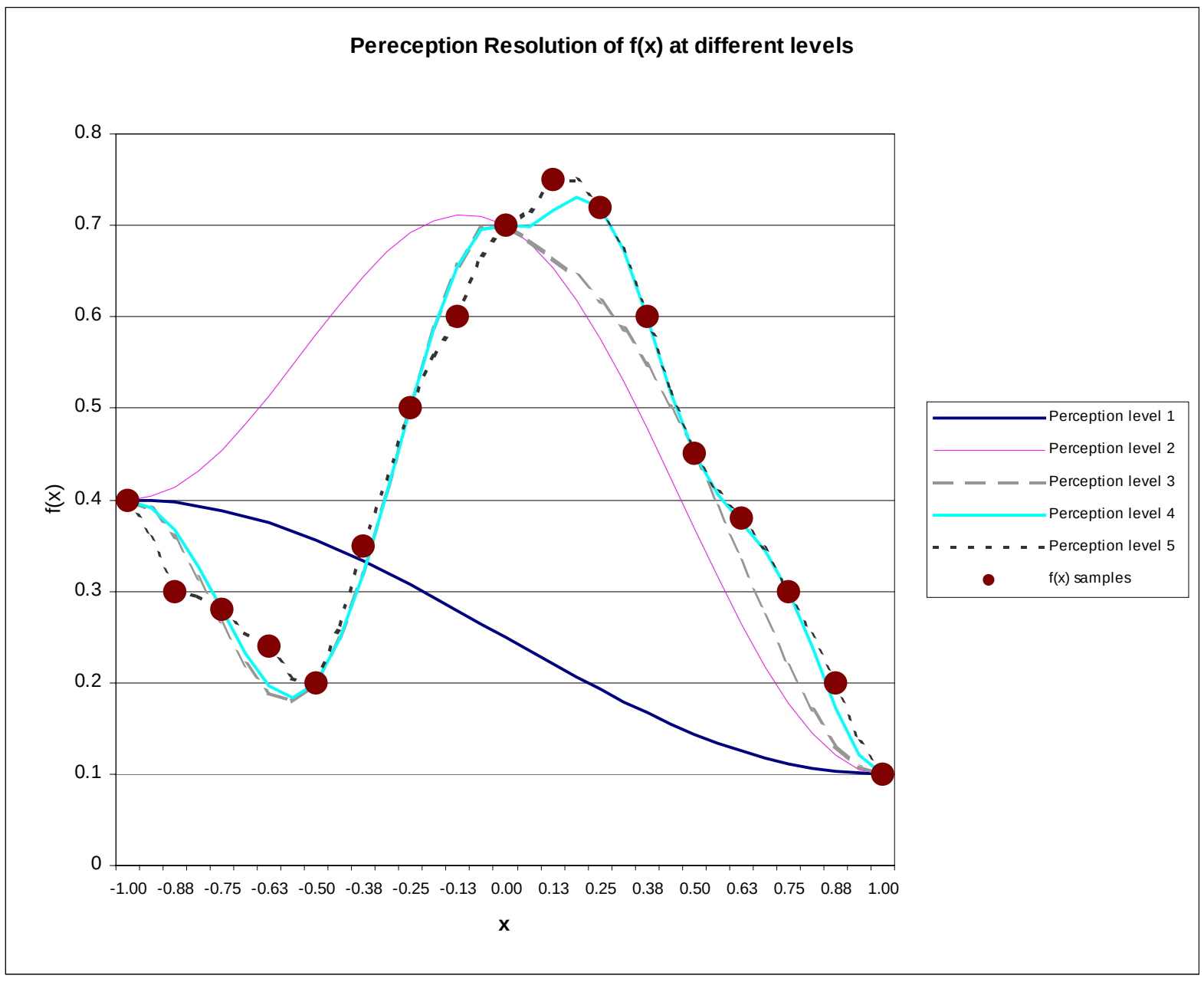

Figure (8.3) : A nonlinear samples approximated and modeled using multi-resolution perception

Additionally, the function of figure (8.2) is represented at the third level of resolution by:

$$f^3(x) = 0.5b^1_{-1} + 0.2b^1_{+1} + 0.35b^2_0 + 0.088b^3_{-\frac{1}{2}} + 0.081b^3_{+\frac{1}{2}} \qquad (8.2)$$

Finally, the function of figure (8.3) at the third level of perception resolution is given by:

$$f^3(x) = 0.4b^1_{-1} + 0.1b^1_{+1} + 0.45b^2_0 - 0.381b^3_{-\frac{1}{2}} + 0.081b^3_{+\frac{1}{2}} \qquad (8.3)$$

On the other hand, figure (8.4) shows the results of learning based on neighborhood learning, i.e. the learning at each boundary is affected by all surrounding samples.

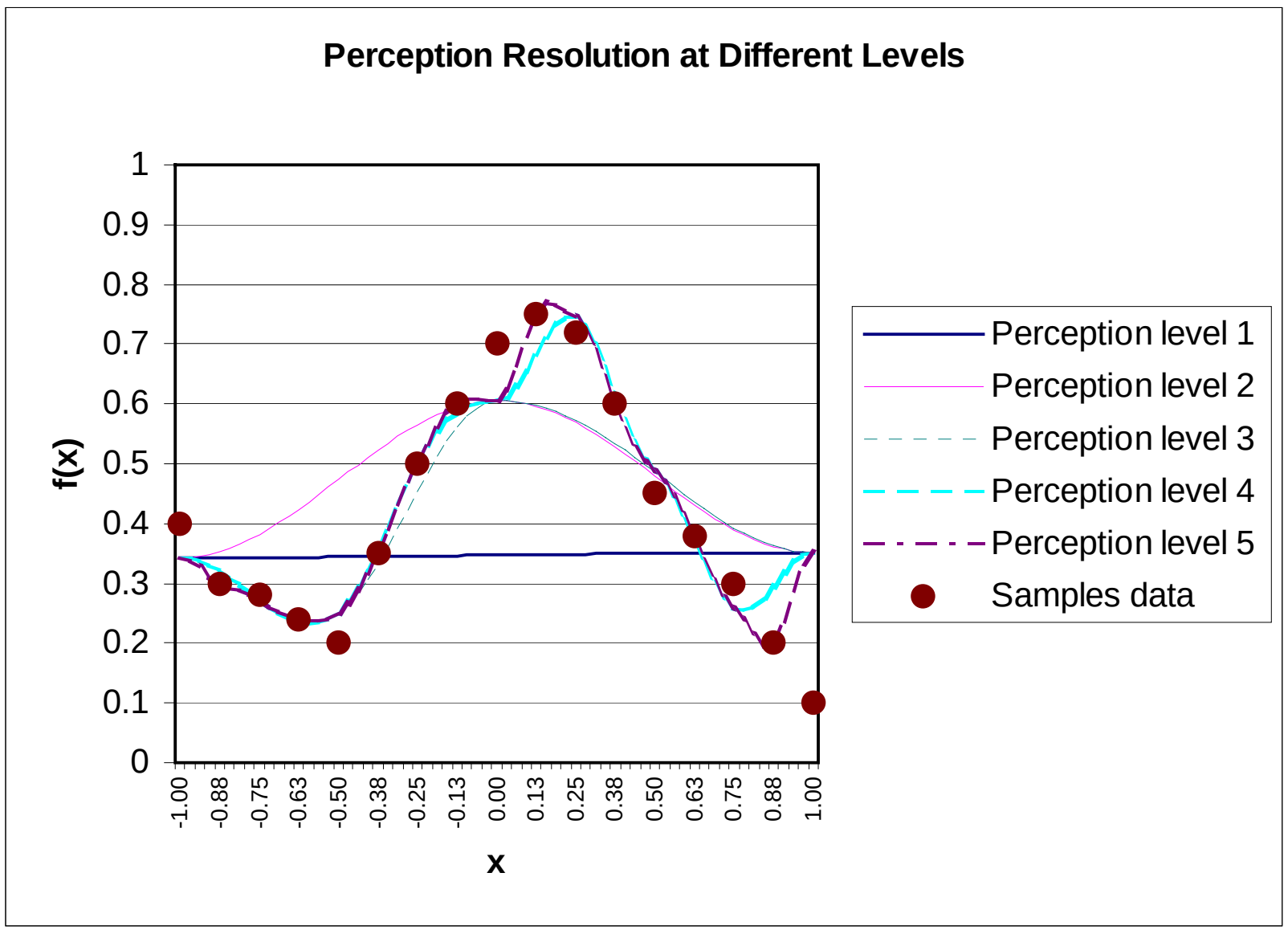

Figure (8.4) : Multi-resolution learning based on neighborhood learning

## 9. Concluding Remarks

The importance of viewing logic as a continuous set form –1 to +1 is that one continuous logic variable is theoretically represented by an infinite of zeros and ones sequence, and hence an infinite amount of information. Therefore we expect one neuron, as a logic-processing unit, to have the property of infinite information-carrying capacity. In this logic, no ambiguity is assumed; the universe of discourse is mapped onto the perception space and can be inversed without any loss of generality.

The introduced model of perception represents a lower level of mental process, i.e. both knowledge acquisition and perception, and it does not model higher level of thinking, planning and so on. Perception automata, based on this approach, is a dynamical system consists of hyper-graph of processing elements; each element can be either excited (taking part in the perception process), or *quiet* (*isolated* or its output is null). This isolation, or the null state, is similar to the high impedance or the third state in logic design.

Based on our view, perception automata does not measure sensory signals directly, rather, it organizes its components in such a way that certain signal potential are distinguished by topological nodes that discriminate them. Learning is accomplished by re-organization of nodes and by adjusting their local weights based only local interaction, therefore, perception automata could be seen as it *geometrize* computation. Symbols are represented internally (topologically) by some nodes, after signals decomposition and aggregation of some features, and this is based on their output. In this model, the equilibrium state of the automata is distributed all over the nodes, and the nodes can be either *excited*, or not excited (output no thing). One advantage of this approach is that both symbols and the ability to change and adapt exist in this model, Luger [15] has mentioned [15] "A theory of how symbols may reduce to patterns in a network and, in turn, influence further adaptation of that network will be an extraordinary contribution. This will support a number of developments, such as integrating network-based perceptual and knowledge intensive reasoning facilities into a single intelligence. " [15].

The way that the perception automata produces an approximation, or exact representation in the perception space, can be viewed as a transformation of these samples into another space. This transformation can be used directly for data compression, it has some similarities with wavelet transform or multi-resolution analysis, but we should note the differences between them, for example, the learning automata is decomposing the function based on top down approach, whereas multi-resolution analysis is a bottom up approach, secondly wavelets assumes the existence of all samples before starting to decompose, whereas perception automata does not requires this condition, thirdly no equal intervals between samples are required; with the addition of any new training data, the *function approximator* simply captures this information, and modifies the structure at different level of resolution and approximation. The proposed learning automata is not ad-hoc solution, such as most approaches of soft computing, rather, it is proved that it can approximate and converge. Therefore, given any training data or samples that represents a relation, this relation or function can be approximated by the perception automata at different levels of perception or resolution. This approximation is expressed in a rigorous mathematical form that enables the designer to take decision about the level of the required approximation or precision.

The difference between our approach and other approaches is clearly wide; the problem with rules-based systems is that they are difficult to be adaptive and to deal with uncertainty and impression. Fuzzy reasoning, irrespective of its mathematical foundations, has also gab between knowledge acquisition and representation. Fuzzy logic is a tool for representing imprecise, ambiguous, and vague information. Its power lies in its ability to perform meaningful and reasonable operations based on concepts that are outside the definitions of conventional Boolean or crisp logic. By introducing elasticity to numbers and symbols, which is so natural to our inability to provide exact measurements. Therefore, although fuzzy logic has elasticity and gradual degree of belonging to symbols, which is represented by fuzzy sets, but it has limited ability to lean and adapt.

On the other hand, current artificial neural networks can not be considered a physical realization of the actual neural networks, and they do not help uncover the internal structure and the structure of the real neural networks, they suffer from long learning time, tendency to overfitting, bad generalization, and non guaranteed convergence. Although symbolic computation and neural networks approaches to intelligence seem to be very different approaches, they share common properties [15]: Firstly, they encode intelligence as computation, secondly they offer a formal model for it, and thirdly they seek physical realization of intelligence.

In a reliable neural network model, we cannot assume that the model can internally calculate, i.e. add, subtract, and perform arithmetic operation explicitly, unless the dynamics of the model are assumed to have certain functionality that can be approximated by such operations. We can not also assume that there exists a centralized learning algorithm (such as in backpropagation net) that adjusts the weights based on some objective function optimization.

Generally, The drawback of neural networks and fuzzy systems is that they can be seen as an ad hoc models and techniques. Both neural networks and fuzzy systems are model free estimators; they do not guess how output functionally depends on input. Traditional A.I. methods are also model free estimators, since they map conditions to actions without declaring a mathematical transfer function that maps from condition space to action space. On the other hand, one favor of neural networks and fuzzy systems is that they are numerical estimators and dynamical systems [1].

## Conclusions

In this report, an innovative approach to model perception was introduced, a logic that can model perceptual information was proposed. Multi-perceptual resolution analysis based on this logic was given. A proposed automata based on this analysis was introduced and a learning mechanism that could converge to whatever required accuracy was provided. Simulation results showed the ability of this automata to learn at different perceptual levels and its convergence speed compared to neural network training process. This report may play a role in our understanding of human cognition and our ability to build machines that can successfully simulate human behavior. This approach combines the advantages of both soft computing and the formal mathematical modeling in the sense that it shares the soft computing approach the features of being model-free approximator and being adaptive. In the mean time, it can be considered a solid mathematical model since it can converges to whatever accuracy we want; secondly it offers a precise description of approximation at each level.